\documentclass[acmtog,screen]{acmart}
\acmSubmissionID{905}

\usepackage{graphicx}
\usepackage[capitalize]{cleveref}

\usepackage{orcidlink}

\usepackage{wrapfig}
\usepackage{multirow}
\usepackage{tabularx}
\usepackage{algorithmic}
\usepackage[ruled]{algorithm2e}
\usepackage{mathtools}
\usepackage{balance}

\usepackage{float}
\usepackage{enumitem}
\usepackage{comment}
\usepackage{pifont}
\usepackage[toc,page]{appendix}

\usepackage{soul}
\usepackage{acronym}
\usepackage{fancybox}
\usepackage{epigraph}
\usepackage{dirtytalk}
\usepackage{bm}
\usepackage{fontawesome}
\usepackage{colortbl}
\usepackage{booktabs}
\usepackage{xspace}

\usepackage{xr}
\usepackage{tikz}
\usepackage{microtype}
\usepackage{transparent}
\usepackage{subfig}
\usepackage{multicol}
\usepackage[htt]{hyphenat}
\usepackage[toc,page]{appendix}

\definecolor{citecolor}{HTML}{0071bc}
\definecolor{frontcolor}{HTML}{325ea5}
\definecolor{backcolor}{HTML}{a58b77}
\definecolor{sidecolor}{HTML}{10768c}
\definecolor{skincolor}{HTML}{dcb7b7}
\definecolor{darkred}{rgb}{0.6, 0.1, 0.05}
\definecolor{DeltaColor}{rgb}{0.039,0.73,0.71}
\definecolor{SigmaColor}{rgb}{0.98,0.45,0.0}
\definecolor{AlphaColor}{rgb}{0,0,0.8}
\definecolor{BetaColor}{rgb}{0.8,0,0.8}
\definecolor{GammaColor}{rgb}{0.514,0.34,0.224}
\definecolor{EpsilonColor}{rgb}{0.353,0.725,0.906}
\definecolor{PurpleColor}{HTML}{8B008B}
\definecolor{BadColor}{HTML}{C0392B}
\definecolor{OrangeColor}{rgb}{0.914,0.541,0.0.141}
\definecolor{GreenColor}{HTML}{00ab41}
\definecolor{RedColor}{rgb}{0.949,0.275, 0.224}
\definecolor{LightCyan}{rgb}{0.88,1,1}
\definecolor{Gray}{gray}{0.85}
\definecolor{LightGray}{gray}{0.70}

\definecolor{greenprior}{HTML}{34a853}
\definecolor{redprior}{HTML}{ea4335}
\definecolor{blueprior}{HTML}{4285f4}

\definecolor{bestcolor}{rgb}{1, 0.5, 0.25}
\definecolor{secondbestcolor}{rgb}{1, 0.8, 0.5}

\makeatletter
\newcommand*{\addFileDependency}[1]{
  \typeout{(#1)}
  \@addtofilelist{#1}
  \IfFileExists{#1}{}{\typeout{No file #1.}}
}
\makeatother

\newlength\savewidth

\usepackage[ruled]{algorithm2e} 

\SetAlFnt{\small}
\SetAlCapFnt{\small} 
\SetAlCapNameFnt{\small}
\SetAlCapHSkip{0pt}

\newcommand{\cmark}{\ding{51}}%
\newcommand{\xmark}{\ding{55}}%

\usepackage{makecell}
\usepackage{multirow}
\usepackage{hyperref}

\acmJournal{TOG}

\author{Xiaoguang Han}
\authornote{Equal Contribution.}
\authornote{Corresponding Author.}
\email{hanxiaoguang@cuhk.edu.cn}
\orcid{0000-0003-0162-3296}
\affiliation{%
  \institution{SSE and FNii, CUHKSZ}
  \country{China}
}

\author{Yushuang Wu}
\authornotemark[1]
\authornote{Work done during internship supervised by Weihao Yuan at Alibaba.}
\email{yushuangwu@link.cuhk.edu.cn}
\orcid{0009-0002-9725-0606}
\affiliation{%
  \institution{FNii and SSE, CUHKSZ}
  \country{China}
}

\author{Luyue Shi}
\authornotemark[1]
\email{117010231@link.cuhk.edu.cn}
\orcid{0009-0002-2012-5014}
\affiliation{%
  \institution{FNii and SSE, CUHKSZ}
  \country{China}
}

\author{Haolin Liu}
\authornotemark[1]
\email{115010192@link.cuhk.edu.cn}
\orcid{0009-0009-4962-9217}
\affiliation{%
  \institution{FNii and SSE, CUHKSZ}
  \country{China}
}

\author{Hongjie Liao}
\email{hongjieliao@link.cuhk.edu.cn}
\orcid{0009-0005-1992-7089}
\affiliation{%
  \institution{FNii and SSE, CUHKSZ}
  \country{China}
}

\author{Lingteng Qiu}
\email{220019047@link.cuhk.edu.cn}
\orcid{0000-0002-3250-0486}
\affiliation{%
  \institution{FNii and SSE, CUHKSZ}
  \country{China}
}

\author{Weihao Yuan}
\email{wyuanaa@connect.ust.hk}
\orcid{0000-0002-1362-3747}
\affiliation{%
  \institution{Alibaba Group}
  \country{China}
}

\author{Xiaodong Gu}
\email{vactor1994@gmail.com}
\orcid{0000-0003-2623-7973}
\affiliation{%
  \institution{Alibaba Group}
  \country{China}
}

\author{Zilong Dong}
\email{zjudzl@qq.com}
\orcid{0000-0002-6833-9102}
\affiliation{%
  \institution{Alibaba Group}
  \country{China}
}

\author{Shuguang Cui}
\email{shuguangcui@cuhk.edu.cn}
\orcid{0000-0003-2608-775X}
\affiliation{%
  \institution{SSE and FNii, CUHKSZ}
  \country{China}
}

\newcommand{\eg}{\textit{e.g.}\xspace}

\def\degree{${}^{\circ}$\xspace}

\begin{document}
\title{MVImgNet2.0: A Larger-scale Dataset of Multi-view Images}

\begin{teaserfigure}
  \centering
  \includegraphics[width=.99\linewidth]{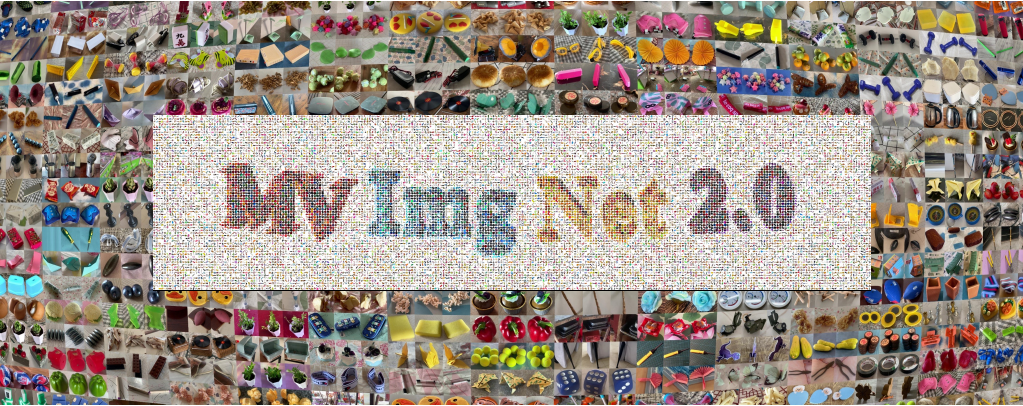}
  \vspace{-0.3cm}
  \caption{We introduce \textbf{MVImgnet2.0}, a larger-scale dataset of multi-view images, which enjoys 3D-aware signals from multi-view consistency. MVImgnet2.0 expands its last version into a total of 520k objects from 515 categories, and also provides higher-quality annotations. The extremely rich geometry and texture information in real-world objects leads to MVImgNet2.0's great potential in supporting large-scale learning in the 3D domain. }
  \label{fig_teaser}
\end{teaserfigure}

\begin{abstract}
MVImgNet is a large-scale dataset that contains multi-view images of $\sim$220k real-world objects in 238 classes. As a counterpart of ImageNet, it introduces 3D visual signals via multi-view shooting, making a soft bridge between 2D and 3D vision. This paper constructs the MVImgNet2.0 dataset that expands MVImgNet into a total of $\sim$520k objects and 515 categories, which derives a 3D dataset with a larger scale that is more comparable to ones in the 2D domain. In addition to the expanded dataset scale and category range, MVImgNet2.0 is of a higher quality than MVImgNet owing to four new features: (i) most shoots capture 360\degree views of the objects, which can support the learning of object reconstruction with completeness; (ii) the segmentation manner is advanced to produce foreground object masks of higher accuracy; (iii) a more powerful structure-from-motion method is adopted to derive the camera pose for each frame of a lower estimation error; (iv) higher-quality dense point clouds are reconstructed via advanced methods for objects captured in 360\degree views, which can serve for downstream applications. Extensive experiments confirm the value of the proposed MVImgNet2.0 in boosting the performance of large 3D reconstruction models. MVImgNet2.0 will be public at \href{https://luyues.github.io/mvimgnet2/}{\textit{\textcolor{magenta}{luyues.github.io/mvimgnet2}}}, including multi-view images of all 520k objects, the reconstructed high-quality point clouds, and data annotation codes, hoping to inspire the broader vision community.

\end{abstract}


\setcopyright{acmlicensed}
\acmJournal{TOG}
\acmYear{2024} \acmVolume{43} \acmNumber{6} \acmArticle{} \acmMonth{12}\acmDOI{10.1145/3687973}


\maketitle


\section{Introduction}
\label{sec:intro}

The field of deep learning has witnessed remarkable advancements, fueled primarily by learning from vast amounts of data~\cite{deng2009imagenet, lin2014mscoco, krishna2017visualgenome, miech2019howto100m}. Learning from large-scale data has proven to be a key driver in scaling up deep learning models to tackle complex understanding or generative tasks, especially for the development of large models in the fields including natural language processing~\cite{achiam2023gpt, thoppilan2022lamda, touvron2023llama}, computer vision~\cite{2024SoraReview, liu2023grounding, kirillov2023segany, ren2024grounded}, and multimodal learning~\cite{liu2024llava, li2023blip, lin2023vila}.

This learning regime also attracts great attention in the field of 3D vision. In spite of the greater difficulty in collecting and labeling 3D data compared with textual or 2D visual data, there are still some efforts contributed to constructing large-scale or high-quality 3D generic datasets~\cite{chang2015shapenet, reizenstein21co3d, yu2023mvimgnet, wu2023omniobject3d, downs2022gso, objaverse, objaverseXL}. Among them, one line of work constructs datasets like ShapeNet~\cite{chang2015shapenet} and Objaverse~\cite{objaverse} composed of synthetic data, which limits the application in real scenarios. Differently, another line of work collects 3D data of real-life objects via scanning or multi-view photogrammetry. However, such datasets like CO3D~\cite{reizenstein21co3d} and GSO~\cite{downs2022gso} are limited in scale and category range until Yu et al. make the first step in constructing a large-scale one, MVImgNet~\cite{yu2023mvimgnet}, consisting of $\sim$220k multi-view images of 238 classes of common objects. The massive multi-view data do not only prove valuable in 2D visual understanding~\cite{yu2023mvimgnet} via learning cross-view consistency, but also support the learning of generic shape priors to benefit 3D reconstruction~\cite{wu2023reconfusion, hong2023lrm, wang2023pf, xu2023dmv3d}.
Considering the larger scale of datasets in the 2D domain, \eg ImageNet~\cite{deng2009imagenet}, containing over 1 million images of 1k categories, MVImgNet is still inferior in scale that may limit its potential to support scaling up 3D learning. Therefore, we propose MVImgNet2.0 that expands MVImgNet to \textbf{twice} its original scale and category range. With a total of \textbf{520k objects} and \textbf{515 categories} that is \textbf{half} the scale of ImageNet, MVImgNet2.0 makes a further step towards a larger real-world 3D dataset with a smaller gap to ones in the 2D domain.   

In addition to the expanded data scale and category range, MVImg-Net2.0 has some other new features in data acquisition and annotation to improve the dataset quality. The biggest difference in data acquisition is that MVImgNet videos usually cover 180\degree views of objects, while most of the videos (230k/300k) collected in MVImgNet2.0 capture \textbf{360\degree object views} to represent a more complete shape. On the other hand, the annotations in MVImgNet2.0 are of higher quality in three aspects: (i) the foreground \textbf{object masks} in each frame are provided of higher accuracy; (ii) the \textbf{camera poses} of each view are estimated of lower error; (iii) the \textbf{dense reconstructions} are advanced to produce object point clouds of higher accuracy and robustness. To get the masks of objects of interest in each video frame, we advance the segmentation method in MVImgNet into a new detection-segmentation-tracking pipeline, which adopts a coarse-to-fine paradigm with temporal information~\cite{liu2023grounding, kirillov2023segany, yang2022deaot} also incorporated to generate accurate object masks finally.
For the camera pose estimation, we apply a more advanced structure-from-motion (SfM) algorithm that refines keypoints and bundles using deep features~\cite{lindenberger2021ppsfm} to compute camera poses of higher accuracy, especially for objects with fewer textures. 
The dense reconstructions in MVImgNet are generated by the multi-view stereo method, which is also advanced in MVImgNet2.0 based on neural surface rendering with multi-resolution 3D hash grids ~\cite{li2023neuralangelo, instantangelo} to improve the reconstruction accuracy and robustness. 

While MVImgNet proves valuable in various visual tasks including radiance field reconstruction, multi-view stereo, and view-consistent image understanding~\cite{yu2023mvimgnet}, our experiments mainly focus on the task of 3D reconstruction. We first demonstrate that the more accurate camera poses estimated in MVImgNet2.0 can better support the per-scene 3D reconstruction. Besides, our experiments pays more attention to the application of generic 3D object reconstruction. We deploy three recent large reconstruction models -- LRM~\cite{hong2023lrm}, LGM~\cite{tang2024lgm} and TriplaneGaussian~\cite{zou2023triplane}, and justify the value of MVImgNet2.0 data in improving their reconstruction quality and generalizability.


The contributions of this work are then summarized as follows:

\begin{itemize}
    \item We propose MVImgNet2.0 that expands MVImgNet to a total of 520k real-life objects and 515 categories, which makes a further step to a larger-scale 3D generic dataset.
    \item MVImgNet2.0 has some new features in both data acquisition and annotation: (i) most of the added videos capture objects in 360\degree views; (ii) frames are annotated with more accurate object masks and (iii) more accurate camera poses; (iv) objects are densely reconstructed with an advanced method.
    \item Extensive experiments validate MVImgNet2.0's value in the task of 3D reconstruction, especially in improving the performance of large 3D reconstruction models.

\end{itemize}

\section{Related Work}
\label{sec:related}

\paragraph{Large-scale datasets.} Expanding the size and breadth of training datasets has proven to be a highly effective strategy for enhancing the performance and robustness of deep learning models. In computer vision, the introduction of large-scale datasets such as ImageNet~\cite{deng2009imagenet} and MS-COCO~\cite{lin2014mscoco} has driven significant advancements across various tasks, including image classification, object detection, and captioning. This trend has persisted, with the diversity and scale of available datasets growing exponentially. Notable examples include image datasets like OpenImages~\cite{kuznetsova2020open} and Visual Genome~\cite{krishna2017visualgenome}, video datasets like Kinetics~\cite{kay2017kinetics} and HowTo100M~\cite{miech2019howto100m}, and multi-modal datasets like Conceptual Captions~\cite{sharma2018conceptual}, YFCC100M~\cite{thomee2016yfcc100m}, and LAION~\cite{schuhmann2022laion}, which support comprehensive vision-language correlations. By increasing the scale and coverage of these datasets, researchers and practitioners have achieved substantial improvements in the capabilities of computer vision and multi-modal systems~\cite{dosovitskiy2020image, liu2023grounding, radford2021learning, jia2021scaling}.

\begin{figure*}[t]
	\centering
	\includegraphics[width=0.95\linewidth]{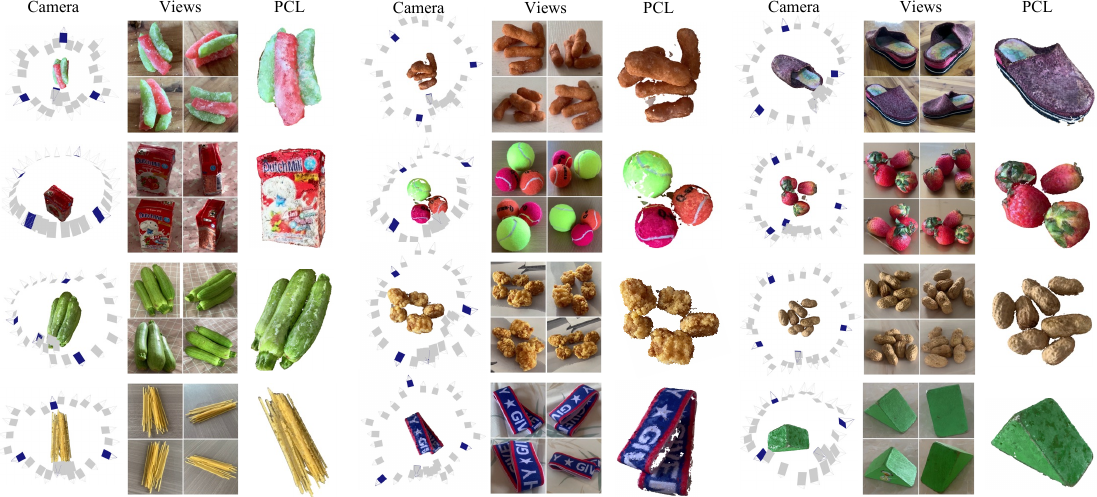}
    \vspace{-0.3cm}
	\caption{MVImgNet2.0 data visualization. Objects in MVImgNet2.0 are in a wide range. For each object, we visualize the estimated camera poses and then sample 4 views to present images (whose corresponding camera poses are highlighted in dark color). We also visualize the point cloud annotations (PCL).  }
	\label{fig:data_vis_main}
 \vspace{-0.3cm}
\end{figure*}

\paragraph{3D datasets.} 3D datasets encompass a broad spectrum, ranging from indoor to outdoor scenes and from human subjects to various objects. This paper primarily focuses on generic object 3D datasets. These datasets can be categorized into two main groups. The first group consists of synthetic 3D objects, such as those found in ShapeNet~\cite{chang2015shapenet}, ModelNet~\cite{wu2015modelnet}, ABO~\cite{collins2022abo}, and Objaverse~\cite{objaverse, objaverseXL}. These datasets offer high-quality computer-aided design (CAD) models as 3D ground truths and can render 2D views in simulation to support 3D reconstruction learning. Though Objaverse collects over 800k CAD models, the quality of models is hard to guarantee, of which only $\sim$170k (21\%) have textures and less are of high quality for training, \eg $\sim$80k (10\%) are used in LGM training~\cite{tang2024lgm}. Besides, the main limitation of such datasets lies in the intrinsic domain gap between synthetic and real objects. Another group focuses on collecting real-world 3D data through scanning or multi-view shooting. Dedicated scanning methods can produce high-quality 3D assets from real-life objects, as seen in ScanObjectNN~\cite{uy2019scanobjectnn}, NAVI~\cite{jampani2023navi}, GSO~\cite{downs2022gso}, and OmniObj3D~\cite{wu2023omniobject3d}, which collect 14k, 8k, and 6k 3D object data, respectively. Given the high cost of scanning, which limits scalability, MVImgNet~\cite{yu2023mvimgnet} collects 220k 3D objects from 238 categories of real-life objects using multi-view shooting, marking a significant step toward constructing large-scale generic 3D datasets comparable to extensive 2D visual datasets. Another related dataset, CO3D~\cite{reizenstein21co3d}, employs a similar data collection method but on a smaller scale, with 19k objects and 38k objects in its new version. The proposed MVImgNet2.0 dataset advances this approach by expanding the scale and category range of MVImgNet to 520k objects in 515 categories, potentially facilitating the learning of large models for 3D understanding and generation.

\paragraph{3D reconstruction.} 3D reconstruction from a single view or multiple views is a challenging task and also an important application for 3D datasets. Significant advancements have been made in single image to 3D reconstruction, starting with early methods focusing on point clouds~\cite{fan2017point,wu2020pq}, voxels~\cite{choy20163d,tulsiani2017multi, chen2019learning}, meshes~\cite{wang2018pixel2mesh,gkioxari2019mesh}, and introducing shape priors like 3D templates~\cite{roth2016adaptive,goel2020shape,kanazawa2018learning,kulkarni2020articulation}, semantics~\cite{li2020self}, and poses~\cite{bogo2016keep,novotny2019c3dpo} has also been extensively researched. With the emerging techniques based on implicit representations like SDFs~\cite{park2019deepsdf,mittal2022autosdf}, occupancy networks~\cite{mescheder2019occupancy}, and NeRF~\cite{mildenhall2020nerf,yu2021pixelnerf,jang2021codenerf,muller2022autorf}, some category-agnostic methods show great generalization potential~\cite{yan2016perspective,niemeyer2020differentiable} but suffer from the lack of fine-grained details~\cite{xu2019disn,yu2021pixelnerf}. The field of 3D reconstruction from multiple views has also been a major focus in computer vision and graphics for decades. Traditional approaches to this task include structure-from-motion (SfM) methods for sparse reconstruction and calibration~\cite{agarwal2011building, pollefeys2004visual, schonberger2016structure, snavely2006photo}, as well as multi-view stereo (MVS) techniques for dense reconstruction~\cite{furukawa2009accurate,pollefeys2008detailed,schonberger2016pixelwise}. More recently, deep learning-based MVS methods have emerged~\cite{cheng2020deep,gu2020cascade,shen2021cfnet,yao2018mvsnet,yao2019recurrent}, providing efficient, high-quality reconstruction through a feed-forward process. More recently, the use of pre-trained image/language models has introduced semantics and multi-view guidance~\cite{radford2021learning,li2022blip,li2023blip,saharia2022photorealistic,rombach2022high} for image-to-3D reconstruction~\cite{liu2023zero,tang2023make,deng2023nerdi,shen2023anything,anciukevivcius2023renderdiffusion,li2023instant3d}. Further, with the emergence of large-scale 3D datasets~\cite{reizenstein21co3d, yu2023mvimgnet, objaverse}, some works explore a purely data-driven approach that learns a large model to reconstruct generic objects in the wild from 3D datasets~\cite{hong2023lrm, wang2023pf, zhang2024gslrm, zou2023triplane, tang2024lgm, xu2024grm}. The proposed dataset has great potential in supporting these large reconstruction models to scale up their 3D reconstruction capability.

\paragraph{Application and impact of MVImgNet.} The MVImgNet dataset~\cite{yu2023mvimgnet}, which provides a vast collection of multi-view images of real objects, has been extensively utilized in a variety of downstream tasks. In addition to fundamental 2D/3D understanding tasks as demonstrated in \cite{chen2024anydoor, ke2024integrating, aubret2024self, lee2024duoduo}, MVImgNet has significantly influenced the field of 3D object reconstruction. It offers a robust 3D prior through its extensive multi-view captures of diverse objects and also enhances the robustness of models in real-world reconstruction scenarios, as shown in \cite{ntavelis2023autodecoding, wu2024reconfusion, hong2023lrm, wang2023pf, jiang2024real3d}. Furthermore, researchers have explored employing MVImgNet data to train or finetune the multi-view diffusion models~\cite{xu2023dmv3d, gao2024cat3d} and video diffusion models~\cite{chen2024v3d, han2024vfusion3d, xie2023dreaminpainter, zuo2024videomv} for high-quality 3D reconstruction. MVImgNet is also applied in some other related tasks, such as scene-level reconstruction and generation~\cite{anciukevicius2024denoising}, generalizable novel view synthesis~\cite{jang2024nvist, zhu2023caesarnerf}, video generation~\cite{he2024cameractrl}, and 3D super resolution~\cite{shen2024supergaussian}. With MVImgNet2.0's larger scale, increased categories, 360-degree capturing, and higher annotation quality, it is anticipated to provide an even stronger 3D prior for 3D reconstruction and to better support other downstream tasks.

\section{Dataset}
\label{sec:dataset}
\begin{figure}[t]
	\centering
    \vspace{0.2cm}
	\includegraphics[width=1\linewidth]{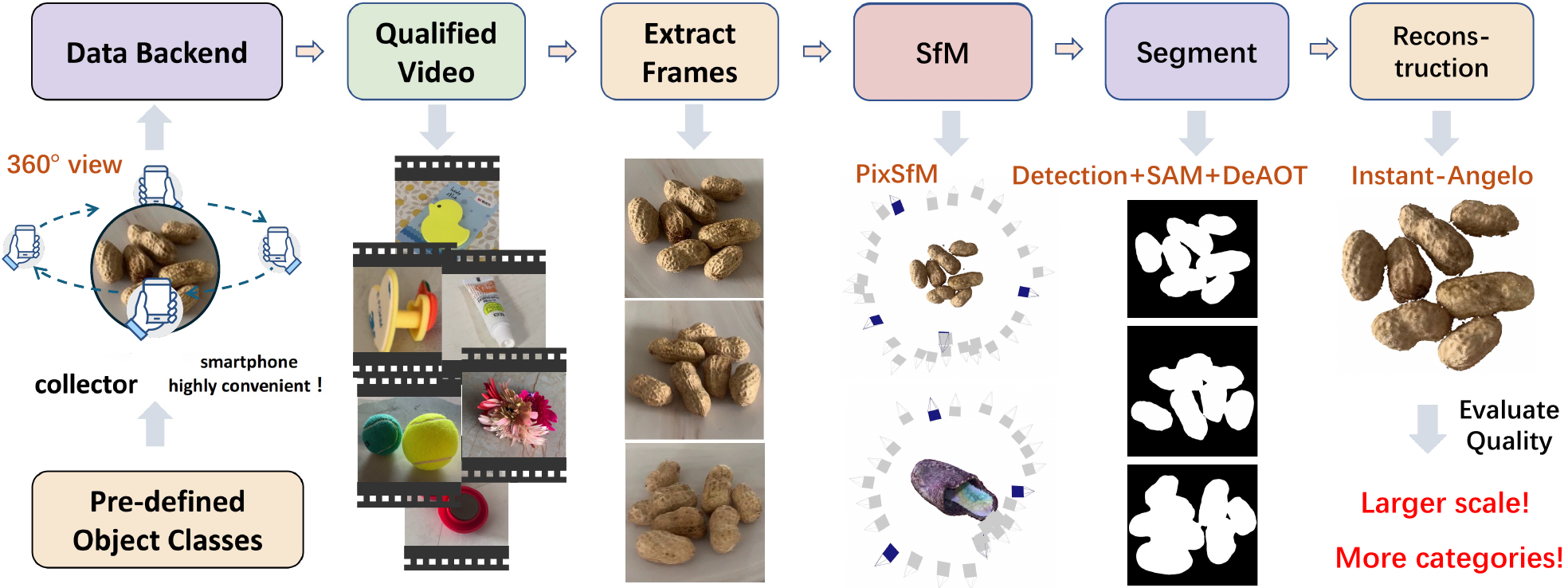}
    \vspace{-0.6cm}
	\caption{The data acquisition and annotation pipeline in MVImgNet2.0. One video is first collected, uploaded, and qualified by collectors and annotators, then we extract frames from the video to conduct annotation including camera pose estimation via PixSfM, then object segmentation via a detection-segmentation-tracking pipeline, and lastly dense point cloud reconstruction via Instant-Angelo. All annotations are qualified by human annotators finally to filter out failure cases. New features in the MVImgNet2.0 pipeline are highlighted in brown or red color. }
	\label{fig:data_pipeline}
 \vspace{-0.4cm}
\end{figure}

MVImgNet2.0 is a large-scale dataset of multi-view images, which is efficiently collected via shooting 360\degree-view videos with phone cameras surrounding objects in the wild. In this section, we introduce the data acquisition and annotation pipeline in constructing MVImgNet2.0, as shown in Fig.~\ref{fig:data_pipeline}, mainly focusing on showing the differences between MVImgNet~\cite{yu2023mvimgnet} and MVImgNet2.0.

\subsection{Raw Data Acquisition}
Similar to MVImgNet, the raw video data is gained through crowd-sourcing. 
We first specify the diverse data categories to collect and the maximum amount for each category.
The categories are chosen following the WordNet~\cite{wordnet} taxonomy and also from the common objects encountered or utilized in human daily life, and the maximum amount is determined by their generality and the complexity involved in capturing them. 
In addition to quantitatively expanding some of MVImgNet's categories with a small number of videos (70 categories), we have also collected 277 new categories to expand MVImgNet's collection of categories. 
Then, we draw up the requirements for the captured videos: (i) The length of each video must be around 10 seconds; (ii) The frames in the video must not be blurred; (iii) The presence proportion of the object in the video frames must be above 15\%; (iv) Each video can only contain one class of principal object; (v) The captured object must be rather ``three-dimensional'' (excluding ones that are too flat and thin, or lacking in depth); (vi) Each video must capture 360\degree view of the object as much as possible. A visualization of camera poses in data collection is shown in Fig.~\ref{fig:data_vis_main}.
After setting up the requirements, similar to MVImgNet, we employ around a thousand normal collectors to take videos and upload them to the backend. Meanwhile, well-trained expert data cleaners are responsible for reviewing each submission and ensuring it fulfills the aforementioned capture requirements. The whole procedure ensures both the diversity and quality of the raw videos.

To sum up, data in MVImgNet2.0 has two main differences from ones in MVImgNet: 1) the data scale and category range are expanded, which allows for learning a more generalizable model; 2) videos are collected by capturing 360\degree view of objects, which allows for learning a better shape prior.


\subsection{Data Annotation}
For each qualified video submission, we exploit a similar data processing procedure as in reconstructing the MVImgNet dataset to conduct semi-automated annotation, as shown in Fig.~\ref{fig:data_pipeline}. At first, around 30 frames are extracted from each video for sparse reconstruction, which derives the estimated camera poses of each view. Then, we generate object masks via segmentation methods for each extracted frame. Finally, given the camera poses and the masks in each view, we conduct dense reconstruction to produce the object point clouds. The main differences in MVImgNet2.0 lie in the advanced approaches to achieve higher-quality annotations, including 1) camera poses, 2) foreground object masks, and 3) point clouds.

\begin{figure}[t]
	\centering
	\includegraphics[width=1\linewidth]{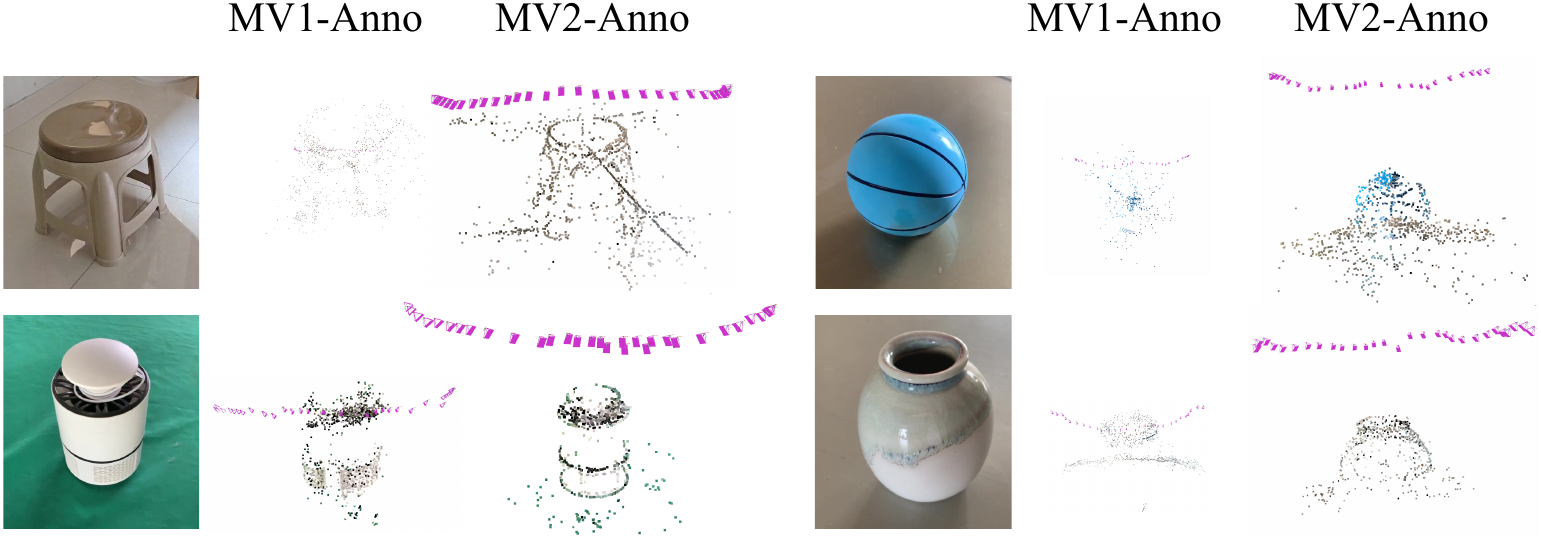}
    \vspace{-0.6cm}
	\caption{The sparse reconstruction results comparison between using the original approach in MVImgNet (MV1-Anno) and using our advanced approach (MV2-Anno) for camera pose estimation (cameras are visualized in purple). }
	\label{fig:data_sfm}
 \vspace{-0.4cm}
\end{figure}

\paragraph{Sparse reconstruction.}
The sparse reconstruction aims to reconstruct the camera intrinsic and extrinsic for each video, by applying the Structure-from-Motion (SfM) algorithm~\cite{schonberger2016sfm} on a series of equal-time-interval chosen frames. In MVImgNet2.0, we apply the Pixel-Perfect Structure-from-Motion (PixSfM) algorithm~\cite{lindenberger2021ppsfm} to obtain the sparse reconstruction results, which can estimate more precise camera parameters via two steps of keypoint and bundle adjustment based on dense features. The sparse reconstruction quality is improved by PixSfM, especially for objects with smooth surfaces and fewer textures, where classical SfM usually fails to produce reasonable estimation, as shown in Fig.~\ref{fig:data_sfm}. 

\begin{figure}[t]
	\centering
    \includegraphics[width=1\linewidth]{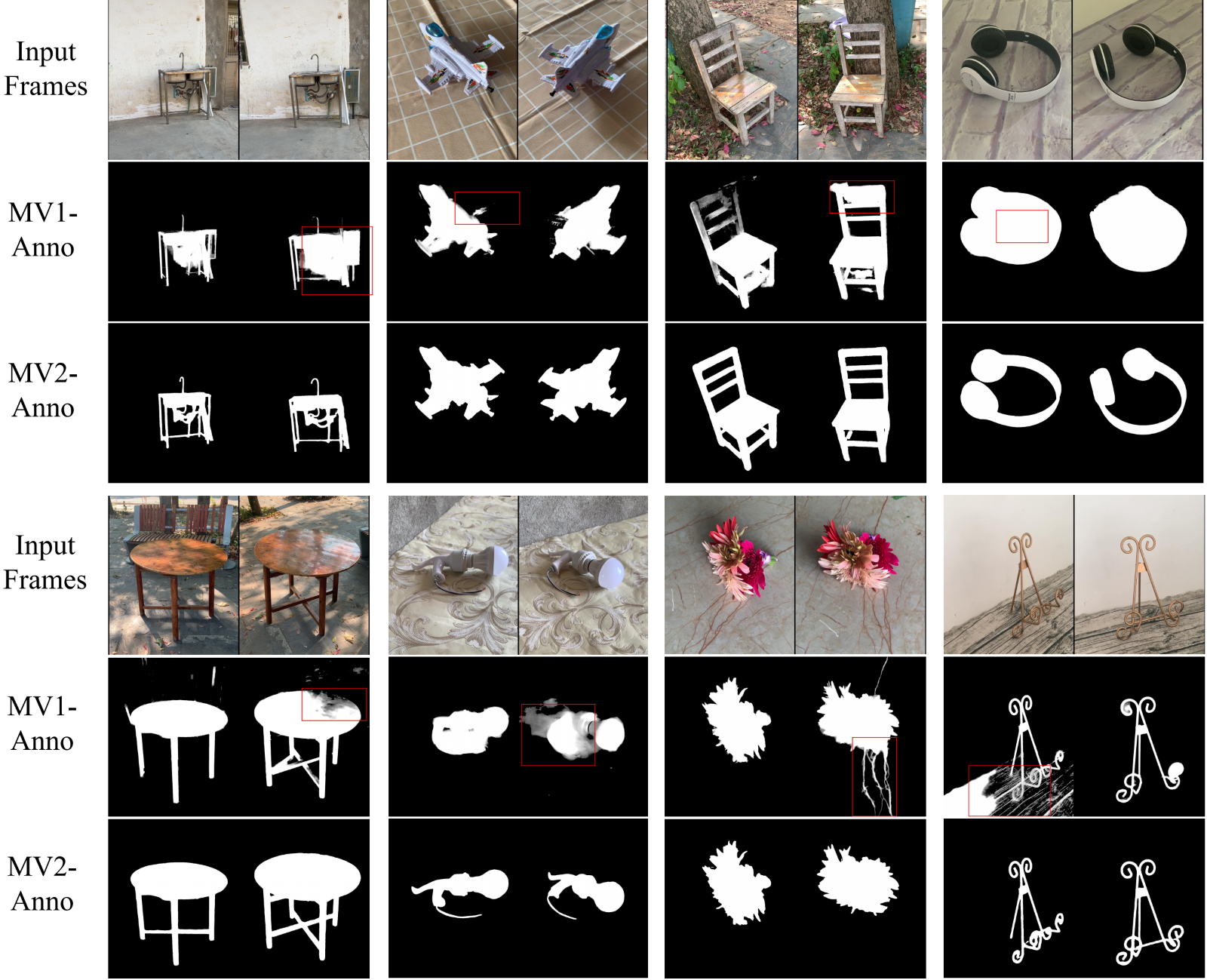}
    \vspace{-0.6cm}
	\caption{The comparison of the foreground object segmentation results between using the original approach in MVImgNet (MV1-Anno) and using our advanced approach (MV2-Anno).}
	\label{fig:data_mask}
 \vspace{-0.3cm}
\end{figure}

\paragraph{Foreground object segmentation}
MVImgNet uses the open-source segmentation tool CarveKit~\cite{carvekit} to generate the foreground object masks, which often results in ambiguous boundaries or incorrect masks, especially for those with a bit complex background.   
To obtain accurate object masks, we apply an advanced detection-segmentation-tracking pipeline, based on an open-set object detector Grounding-DINO~\cite{liu2023grounding}, a segmentation tool Segment-Anything (SAM)~\cite{kirillov2023segany}~\cite{ke2024segment}, and a video object tracker DeAOT~\cite{yang2022deaot}. Given a sequence of video frames, we first apply Grounding-DINO to generate bounding box (bbox) candidates of foreground objects, where the category name is used as the text prompt. With the detection results giving coarse indications, we then apply SAM to generate an object mask for each bbox by using the image as input and the bbox as the prompt. We also initially filter out masks that may be inaccurate, according to their size, distance from the image boundary, number of connected components, etc. Finally, we further take the temporal information, the relation between neighboring frames, into consideration, where one mask is selected as the input of the video object tracker DeAOT to generate the final accurate masks. In addition, to obtain more precise object masks, we also manually check some results of each category and adjust the segmentation pipeline for some categories. We visualize some segmentation results in Fig.~\ref{fig:data_mask} to show the improved mask quality in MVImgNet2.0. A quantitative comparison between the segmenting performance by the original and advanced approach is presented in the SupMat's Tab.~\ref{tab:supp_exp_mask}. 

\begin{figure}[t]
	\centering
	\includegraphics[width=1\linewidth]{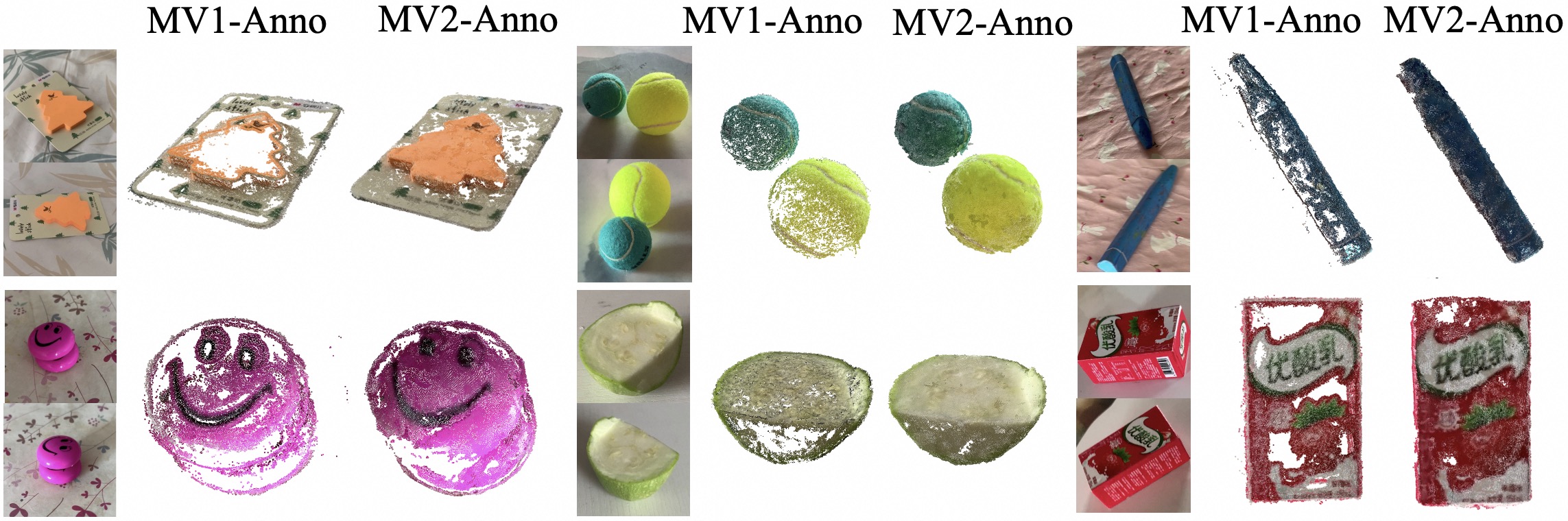}
    \vspace{-0.6cm}
	\caption{The dense point cloud reconstruction results comparison between using the original approach in MVImgNet (MV1-Anno) and using our advanced approach (MV2-Anno).}
	\label{fig:data_mvs}
 \vspace{-0.4cm}
\end{figure}

\paragraph{Dense reconstruction.} 
Different from MVImgNet which employs multi-view stereo (MVS)~\cite{schonberger2016pixelwise} of COLMAP to generate the densely reconstructed point cloud, we advance the point cloud reconstruction approach based on a neural surface reconstruction method, Neural-Angelo~\cite{li2023neuralangelo}. It incorporates multi-resolution hash encoding into neural SDF representations that allows for high-fidelity dense 3D reconstruction. In implementation, we adopt the open-source Instant-Angelo project~\cite{instantangelo} to achieve fast point cloud reconstruction. 
Given the Neural-Angelo reconstruction outputs, similar to MVImgNet, we also manually clean the point clouds to delete the objects with obvious noisy, extremely sparse reconstructions, or backgrounds. Example comparisons between point clouds produced by the approach used in MVImgNet and MVImgNet2.0 are visualized in Fig.~\ref{fig:data_mvs}, which demonstrates that the advanced method used in MVImgNet2.0 usually leads to more accurate and complete reconstructions.

\subsection{Dataset Statistics}

\begin{figure*}[t]
	\centering
        \includegraphics[width=0.94\linewidth]{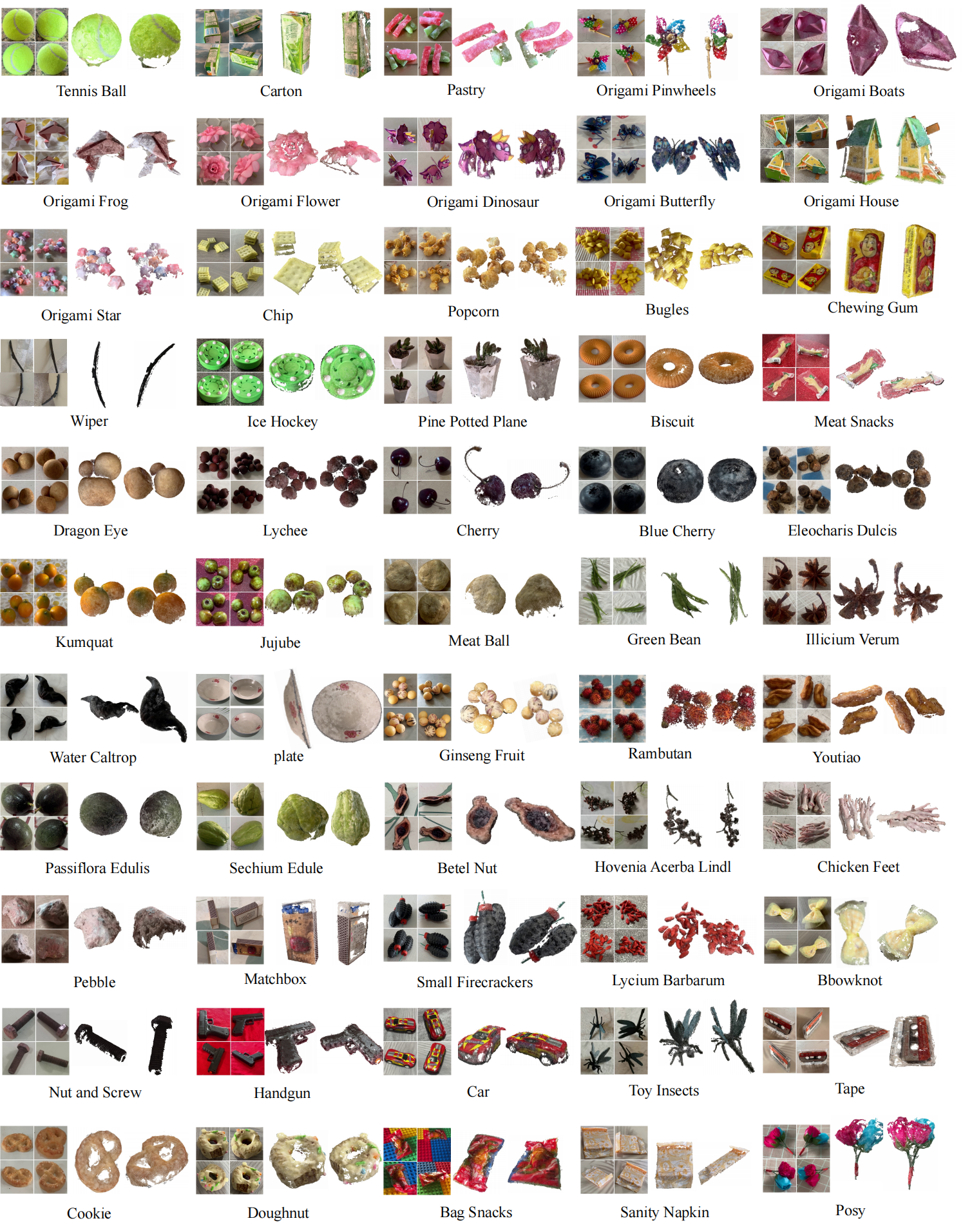}
    \vspace{-0.4cm}
	\caption{Visualization of more reconstructed point cloud annotations from different object categories in MVImgNet2.0. }
	\label{fig:supp_pcl}
 \vspace{-0.4cm}
\end{figure*}

Tab.~\ref{tab:dataset_comp} shows the statistics of MVImgNet2.0 and other alternatives and Fig.~\ref{fig:data_vis_main} and Fig.~\ref{fig:supp_pcl} shows some samples of MVImgNet2.0. In summary, MVImgNet2.0 includes 300k videos with 9 million frames and 347 object classes, of which 277 are new categories not covered by MVImgNet, and the annotations comprehensively cover object masks, camera pose parameters, and point clouds. The categories are organized in a taxonomic manner in SupMat's Fig.~\ref{fig:supp_taxonomy}, and we recommend our project page for a more detailed display. In addition, we also give more detailed per-category statistics in SupMat's Fig.~\ref{fig:supp_cls_amount}. 
With the construction of MVImgNet2.0, the total statistics of MVImgNet datasets reach 520k objects in 515 categories, which is closer to the scale of 2D large-scale datasets, \eg ImageNet with $\sim$1 million data in 1000 categories.



\begin{table}[t]
    \centering
    \caption{\textbf{Comparison} between MVImgNet2.0 and related datasets. “pcl” denotes point clouds. $^*$Note that only parts of data in CO3D ($\sim$30\%) and MVImgNet ($\sim$40\%) are annotated with point cloud GT. }
    \vspace{-3mm}
    \resizebox{\linewidth}{!}{
    \begin{tabular}{l|c|c|c|c|c}
    \toprule
    Dataset & Real &\# of objects & \# of classes & Multi-view & 3D-GT\\
    \midrule
    ShapeNet ~\cite{chang2015shapenet} &\xmark & 51k & 55& render& CAD model\\
    ModelNet~\cite{wu2015modelnet}  & \xmark& 12k & 40 & render& CAD model\\
    ScanObjectNN~\cite{uy2019scanobjectnn} & \cmark& 14k & 15 & limited & pcl\\
    CO3D~\cite{reizenstein21co3d} & \cmark& 19k & 50 & 360$^{\circ}$ views & pcl$^*$\\
    GSO~\cite{downs2022gso} &\cmark & 1k & 17 & 360$^{\circ}$ views & RGB-D scan\\
    ABO~\cite{downs2022gso} &\xmark & 8k & 63 & render & CAD model\\
    Objaverse~\cite{objaverse} & \xmark & 818k & 21k & render & CAD model\\
    OmniObj3D~\cite{wu2023omniobject3d} & \cmark& 6k & 190 & 360$^{\circ}$ views & RGB-D scan\\
    MVImgNet1.0~\cite{yu2023mvimgnet} & \cmark& 220k & 238 & 180$^{\circ}$ views & ~~pcl$^*$\\
    \rowcolor{gray!20}
    MVImgNet2.0 & \cmark& 300k & 347 & 180$^{\circ}$/360$^{\circ}$ views & pcl\\
    \rowcolor{gray!20}
    MVImgNet1.0+2.0 & \cmark& 520k & 515 & 180$^{\circ}$/360$^{\circ}$ views & pcl\\
    \bottomrule
    \end{tabular}
    }
    \label{tab:dataset_comp}
    \vspace{-0.3cm}
\end{table}
\section{Experiments}
\label{sec:experiments}



This section aims to validate the value of the proposed MVImgNet2.0 in the application of 3D reconstruction. We first introduce the experiment setup, then conduct per-scene 3D reconstruction to validate the value of camera pose annotations with higher accuracy, and finally we pay the main focus on justifying the value of new features in MVImgNet2.0 in improving the performance of large reconstruction models, including the larger data scale, the expanded category range, the 360\degree-view videos, and the higher-quality annotations. 

\subsection{Experiment Setup}
\paragraph{Datasets.} 
We adopt three datasets in experiments, including the synthetic dataset Objaverse~\cite{objaverse}, the original data in MVImgNet~\cite{yu2023mvimgnet} (MV1-Data), and the newly added data in MVImgNet2.0 (MV2-Data). The training set includes multiple views captured by videos with estimated camera poses or obtained via rendering the synthetic models from random camera poses. Each object has over 30 views to support training. The test set consists of 1k data sampled from 20 held-out categories in MVImgNet2.0 that are unseen in training. Each test sample contains one or more views with estimated camera poses as input and 8 posed novel-view images with the resolution of 512$\times$512 as the ground truths. Besides, we also provide high-quality dense point cloud reconstructions with manual cleaning by annotators for each sample in the test set as their shape ground truths. 

\paragraph{Evaluation metrics.} As MVImgNet2.0 can provide 2D multi-view ground truths, we mainly employ PSNR/SSIM (higher is better) and LPIPS~\cite{zhang2018unreasonable} (lower is better) as the evaluation metrics to measure the reconstruction quality in projected views. In the evaluation of category-agnostic reconstruction, all backgrounds of test views are masked out as in the training set to focus on the reconstruction accuracy of foreground objects, but preserved in per-scene reconstruction experiments. As TriplaneGaussian also outputs the reconstructed point cloud shape, we also employ Chamfer Distance (CD) as the measurement of the reconstructed shape quality.

\paragraph{Baselines.} Our baselines attempt to cover a wide range of reconstruction models. For per-scene 3D reconstruction, we utilize two baselines, Instant-NGP (INGP)~\cite{muller2022instantngp} and 3D Gaussian Splatting (3DGS)~\cite{kerbl20233dgs}, which are based on the technique of neural radiance field (NeRF) and Gaussian splatting~\cite{kerbl20233dgs}, respectively. Furthermore, we adopt three large reconstruction models as the baselines of category-agnostic 3D reconstructions, which conduct data-driven shape learning from large-scale 3D data for reconstructing generic real-life objects: 
(i) Large Multi-View Gaussian Model (LGM)~\cite{tang2024lgm} that can serve for multi-view reconstruction based on the 3D Gaussian representation; 
(ii) Large Reconstruction Model (LRM)~\cite{hong2023lrm} that addresses single-view reconstruction based on the NeRF representation; 
(iii) Triplane Meets Gaussian Splatting (TriplaneGaussian)~\cite{zou2023triplane} that addresses single-view reconstruction requires point cloud supervision in training. 

\paragraph{Implementation details.} The implementations of INGP, LGM, and TriplaneGaussian follow the official codes, while 3DGS and LRM are implemented following two open-source projects, GauStudio~\cite{ye2024gaustudio} and OpenLRM~\cite{openlrm}. We implement the base version of each 3D reconstruction model if not specified. All hyper-parameters and training strategies follow the recommended or default setting in the papers or released projects unless specified. We use NVIDIA A100 GPUs to train these baselines. In the experiments for investigating the training data factors, we apply LGM-tiny, following the ablation study setting in the LGM paper~\cite{tang2024lgm}. Note that LGM, in the original paper, is adopted to reconstruct from four views generated by multi-view diffusion models to address the task of image-to-3D or text-to-3D generation, in our experiments we directly feed four object views into LGM to perform multi-view reconstruction.

\subsection{Per-scene 3D Reconstruction}
We begin our investigation with per-scene 3D reconstruction of object-centric scenarios, utilizing two baseline methods: INGP and 3DGS.
We randomly choose 50 objects from 25 categories (2 scenes for each category) to conduct experiments. Among a total of 30 views for each object, we randomly select 20 of them for optimizing the parameters in INGP~\cite{muller2022instantngp} and 3DGS~\cite{kerbl20233dgs}, and use the remaining 10 views for evaluation. 
We design two sets of controlled experiments, each differentiated by a single variable: the camera poses utilized during training.
For the first group, the camera poses are estimated via the annotation approach used in MVImgNet (MV1-Anno), while the second group employs the advanced approach in MVImgNet2.0 (MV2-Anno). 


\paragraph{Results.} 
We compute the results averaged across the selected 50 scenes. 
The quantitative results are shown in Tab.~\ref{tab:exp_camera}, where both INGP and 3DGS can get a more accurate reconstruction using the camera poses estimated via the advanced approach in MVImg-Net2.0. By using the camera poses estimated by the advanced approach in MVImgNet2.0, INGP can achieve a higher average PSNR by $\sim$1.1dB, and 3DGS can achieve a significant improvement of $\sim$5.8dB in PSNR.
It validates the higher quality of camera poses estimated in MVImgNet2.0. These results also indicate that MVImgNet2.0 can better support the learning-based reconstruction methods in the task of per-scene reconstruction or novel view synthesis. 
\begin{table}[t]
	\centering
     \caption{\textbf{Per-scene 3D reconstruction quality} comparison when using the estimated camera poses by the annotation manners in MVImgNet (MV1-Anno) and MVImgNet (MV2-Anno) for training. Two baselines are used: Instant-NGP (INGP) and 3D Gaussian splatting (3DGS). } 
	\vspace{-0.3cm}
    \resizebox{1.0\linewidth}{!}{
        \begin{tabular}{l|cc|ccc}
        \toprule
        Baseline & MV1-Anno & MV2-Anno & PSNR$\uparrow$ & SSIM$\uparrow$ &  LIPIS$\downarrow$  \\ 
        \midrule
        \multirow{2}{*}{INGP~\cite{muller2022instantngp}} 
        & \cmark &  & 36.05 & 0.980 & 0.023 \\
        &  & \cmark & \textbf{37.17} & \textbf{0.984} & \textbf{0.018} \\ 
        \midrule
        \multirow{2}{*}{3DGS~\cite{huang20242d}} 
        & \cmark &  & 36.44 & 0.982 & 0.027 \\
        &  & \cmark & \textbf{42.19} & \textbf{0.991} & \textbf{0.015} \\
        \bottomrule
        \end{tabular}
    }  
\vspace{-0.1cm}
\label{tab:exp_camera} 
\end{table}

\subsection{Category-agnostic 3D Reconstruction}
On the task of category-agnostic 3D reconstruction, we perform experiments to evaluate the performance of large reconstruction models trained on different kinds of data to validate the value of MVImgNet2.0. We apply an LGM~\cite{tang2024lgm} for the task of multi-view reconstruction, and an LRM~\cite{hong2023lrm} and a TriplaneGaussian~\cite{zou2023triplane} to address the single-view reconstruction. By using different kinds of multi-view data in training, we investigate their effect on the learning of large reconstruction models. Besides, we also use different kinds of point cloud supervision in training TriplaneGaussian to further validate the value of the point cloud annotations in MVImgNet2.0. Finally, we deeply investigate the factor of data scale, category range, and view range in training data by evaluating their effects on the performance of a tiny LGM (LGM-tiny).

\begin{figure}[t]
	\centering
	\includegraphics[width=1.0\linewidth]{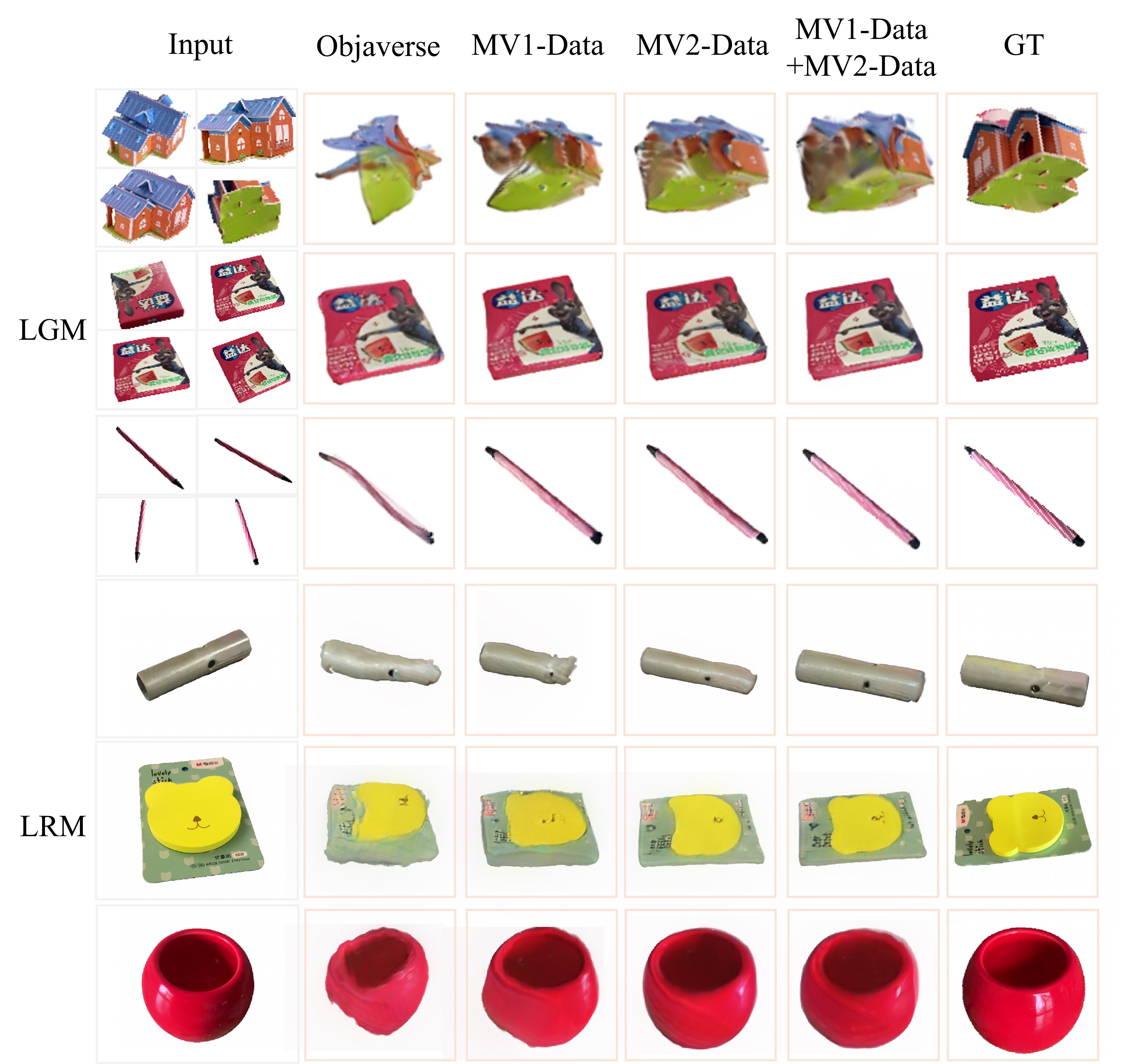}
    \vspace{-0.6cm}
	\caption{Qualitative results of LGM and LRM trained on different data from Objaverse, MV1-Data, and MV2-Data.}
	\label{fig:exp_lrmlgm}
 \vspace{-0.3cm}
\end{figure}

\paragraph{Experiments on LGM and LRM} 
We train LGM and LRM on three kinds of data: (i) synthetic data from Objaverse~\cite{objaverse}; (ii) real data from MVImgNet~\cite{yu2023mvimgnet} (MV1-Data); and (iii) added data in MVImgNet2.0 (MV2-Data). In training, we use 4 randomly selected views as input, and 20 views as supervision to train an LGM-base model, and use 1 randomly selected input view to train an LRM-base model. As shown in Tab.~\ref{tab:exp_lrmlgm}, the quantitative results of LGM and LRM consistently demonstrate: 1) Using real data for training can derive a stronger large reconstruction model when applied to real-world objects. Compared with using Objaverse data for training, using MV1-Data for training can achieve a higher PSNR by $\sim$1.9dB for LGM and by $\sim$0.7dB for LRM; 2) Though with a close scale, using MV2-Data only (300k objects) can lead to a higher reconstruction quality ($\sim$0.5dB$\uparrow$ for both LRM and LGM) compared with using MV1-Data only (220k objects), thanks to the higher quality of MV2-Data; 3) Further incorporating MV2-Data into MV1-Data in training can further bring performance gains, \textit{i.e.}, $\sim$0.4dB and $\sim$0.3dB for LGM and LRM, respectively, which additionally confirm the value of MVImgNet2.0 in benefiting data-driven 3D shape learning. We also provide some qualitative results in Fig.~\ref{fig:exp_lrmlgm} to visualize the differences in reconstruction quality.

\begin{figure}[t]
	\centering
	\includegraphics[width=1\linewidth]{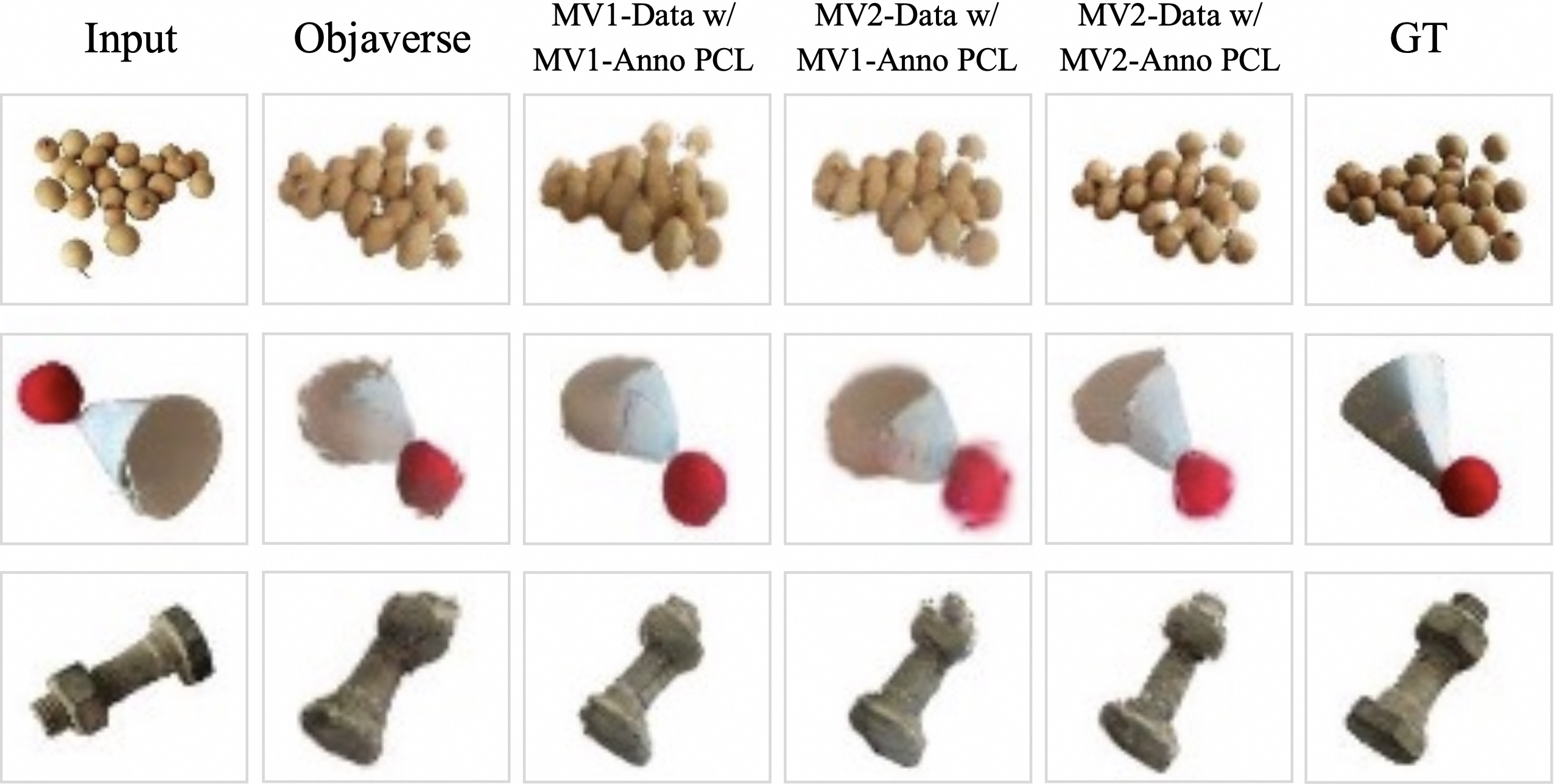}
    \vspace{-0.6cm}
	\caption{Qualitative results of TriplaneGaussian trained on different view data and point cloud supervision.}
	\label{fig:exp_triplane}
 \vspace{-0.4cm}
\end{figure}

\paragraph{Experiments on TriplaneGaussian} We train TriplaneGaussian with different multi-view data and point cloud supervision to further investigate the value of 360\degree-view data and the proposed point cloud annotations in MVImgNet2.0. As shown in Tab.~\ref{tab:exp_triplane}, training on the Objaverse synthetic data results in poor generalization on real data, getting the lowest PSNR in rendering quality. However, since Objaverse can provide perfect point cloud supervision sampled from the ground-truth object surface, training on Objaverse can lead to an acceptable level of shape quality. As MVImgNet only collects 180\degree views for each object, training on MV1-Data with incomplete point cloud supervision leads to low quality in both 2D rendering and 3D shape. Training on 360\degree views (MV2-Data) but with point cloud supervision obtained via the annotation approach in MVImgnet (MV1-Anno) can bring better reconstruction results in rendering quality but poor performance in shape quality. Further using higher-quality point cloud supervision (MV2-Anno) can lead to improvements in the overall reconstruction quality. To sum up, the model trained on 360\degree real-world data (MV2-Data) with more complete point cloud supervision (MV2-Anno) can achieve a higher rendering quality by $\sim$1.0dB in PSNR and also a lower Chamfer distance by 4$\times10^{-4}$ than the one trained on Objaverse. Qualitative results in Fig.~\ref{fig:exp_triplane} also confirm our claim. Thus, the experiment results validate the value of 360\degree-view data and the higher quality of point cloud annotations provided in MVImgNet2.0. 

\paragraph{Training data factors.} We further investigate the effects of three factors in training data on training large reconstruction models: data scale, category range, and view range. The baseline model is based on an LGM-tiny for multi-view reconstruction, and the image resolution for training and testing is 256$\times$256 for efficiency. In the first group of experiments, we basically use 40k MV1-Data and progressively increase the number of MV2-Data from 20k to 140k added for training. As shown in Fig.~\ref{fig:exp_factors}(a), LGM achieves increasingly better performance with the added MV2-Data grows in scale.
In the second group, we hold the added MV2-Data volume constant as 100k but increase the number of categories covered in these data. As the category range expanded, LGM can also achieve iterative performance gains, as shown in Fig.~\ref{fig:exp_factors}(b). 
Next, we control the ratio of 360\degree-view data in a total of 100k MV2-Data added, and find that a higher ratio of 360\degree-view data is positive to the final reconstruction quality of LGM (see Fig.~\ref{fig:exp_factors}(c)). 
These results indicate that training on data of a larger scale, covering richer categories, and capturing a wider view range are important to improve the performance of large reconstruction models. Finally, we use a constant scale of 100k data (MV1-Data + MV2-Data) in total for training, but gradually increase the ratio of MV2-Data from 0\% to 100\%. It leads to a gap of $\sim$0.5dB between using 0\% and 100\% of MV2-Data, as shown in Fig~\ref{fig:exp_factors}(d), which demonstrates that with higher quality, MV2-Data is of larger value than MV1-Data for the learning of large reconstruction models. Some qualitative results of LGM-tiny are included in SupMat's Fig.~\ref{fig:supp_fig_lgm}.

\begin{table}[t]
	\centering
    \caption{\textbf{Generalizable 3D reconstruction quality} comparison when using different set of multi-view data for training. Note that when neither MVImgNet data (MV1-Data) nor MVImgNet2.0 data (MV2-Data) is used, the baseline model is trained on synthetic data from the Objaverse dataset. Two baselines are used: LRM and LGM, for single-view and multi-view reconstruction, respectively. }
    \vspace{-0.3cm}
    \resizebox{0.95\linewidth}{!}{
        \begin{tabular}{l|cc|ccc}
        \toprule
        Baseline & MV1-Data & MV2-Data & PSNR$\uparrow$ & SSIM$\uparrow$ &  LIPIS$\downarrow$  \\ 
        \midrule
        \multirow{4}{*}{LGM~\cite{tang2024lgm}}
        &        &        & 23.59 & 0.932 & 0.050 \\ 
        & \cmark &        & 25.49 & 0.951 & 0.035 \\ 
        &        & \cmark & 26.01 & 0.953 & 0.034 \\ 
        & \cmark & \cmark & \textbf{26.41 }& \textbf{0.956} & \textbf{0.032} \\ 
        \bottomrule
        \multirow{4}{*}{LRM~\cite{hong2023lrm}}
        &        &        & 20.76 & 0.929 & 0.065 \\ 
        & \cmark &        & 21.42 & 0.932 & 0.059 \\ 
        &        & \cmark & 21.97 & 0.935 & 0.056 \\ 
        & \cmark & \cmark & \textbf{22.27} & \textbf{0.936} & \textbf{0.033} \\ 
        \bottomrule
        \end{tabular}
    }   
	\label{tab:exp_lrmlgm} 
\vspace{-0.2cm}
\end{table}

\begin{table}[t]
	\centering
     \caption{\textbf{Generalizable 3D reconstruction quality} comparison when using different multi-view data and point cloud supervision to train a TriplaneGaussian model. The first row uses only Objaverse synthetic data in training, while the second row uses MVImgNet real data (MV1-Data) for training with point clouds obtained via the annotation manner in MVImgNet (MV1-Anno). The last two rows use added MVImgNet2.0 data (MV2-Data) for training, with point cloud supervision (PCL super.) are obtained via different annotation approaches used by MVImgnet (MV1-Anno) and MVImgNet2.0 (MV2-Anno). } 
     \vspace{-0.3cm}
    \resizebox{1.0\linewidth}{!}{
        \begin{tabular}{ccc|cccc}
        \toprule
        MV1-Data & MV2-Data & PCL super. & PSNR$\uparrow$ & SSIM$\uparrow$ &  LIPIS$\downarrow$  &  CD$\downarrow$ ($\times10^{-2}$)  \\ 
        \midrule
               &        & Objaverse  & 21.79 & 0.923 & 0.062 & 0.40 \\
        \cmark &        & MV1-Anno   & 22.43 & 0.924 & 0.060 & 0.82 \\
               & \cmark & MV1-Anno   & 22.51 & 0.928 & 0.056 & 0.89 \\ 
               & \cmark & MV2-Anno   & \textbf{22.77} & \textbf{0.929} & \textbf{0.053} & \textbf{0.36} \\ 
        \bottomrule
        \end{tabular}
    }  
	
	\label{tab:exp_triplane} 
\vspace{-0.3cm}
\end{table}

\newcommand{\gtfigwidth}{0.15\linewidth}
\newcommand{\resultswidth}{0.48\linewidth}

\newcommand{\insertimg}[1]{
  \makecell{
  \includegraphics[width=\resultswidth]{#1} \\
  }
}

\begin{figure}[t]
	\centering
	\begin{tabular}{ccc}
		\insertimg{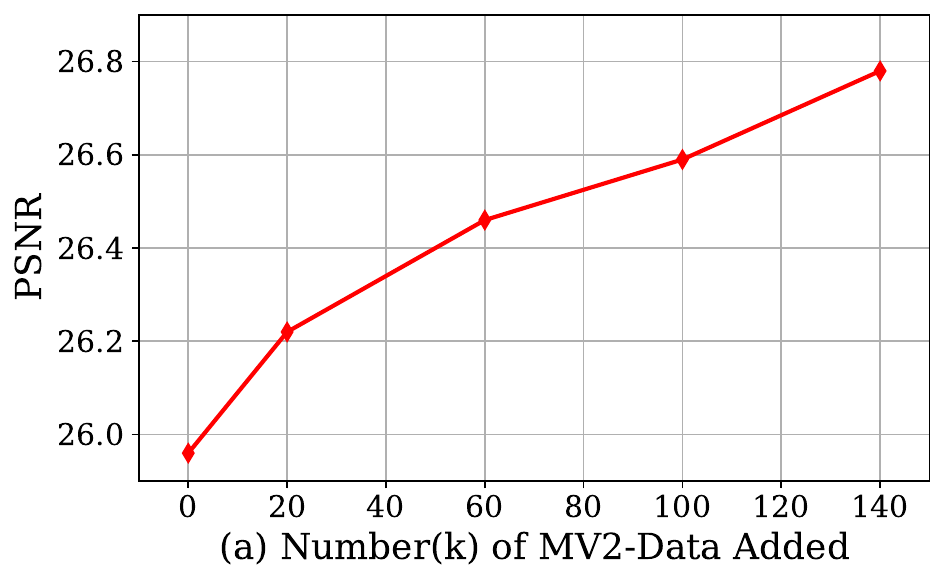}
		\insertimg{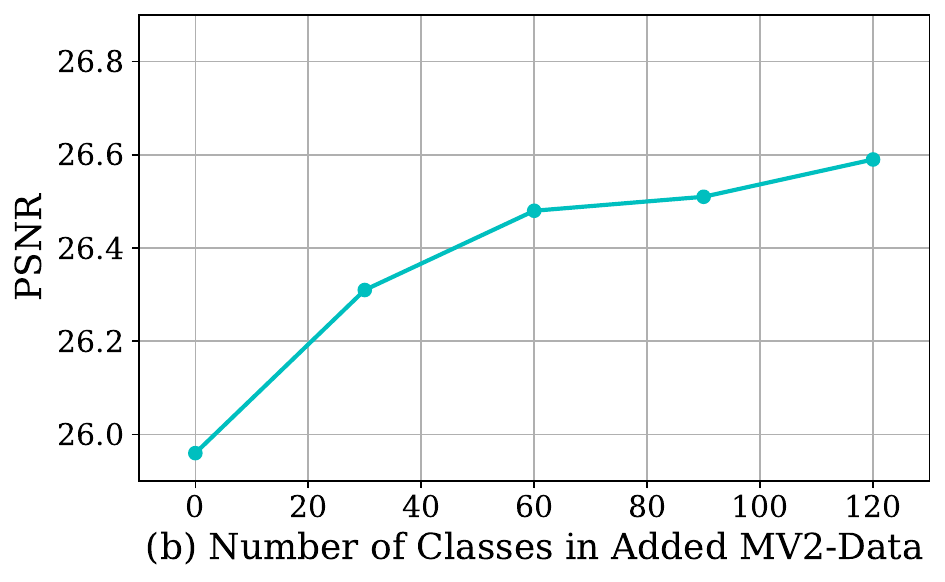} \\
        \insertimg{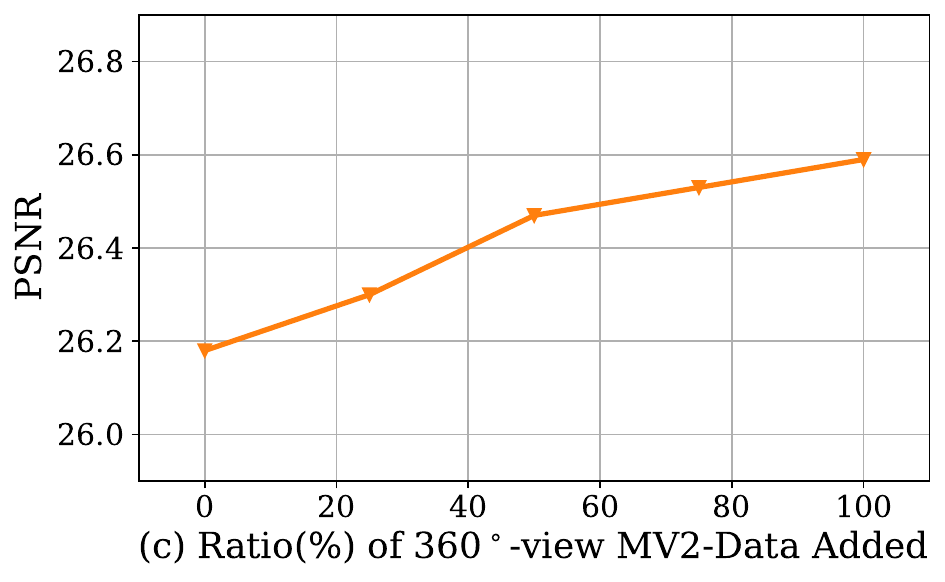} 
		\insertimg{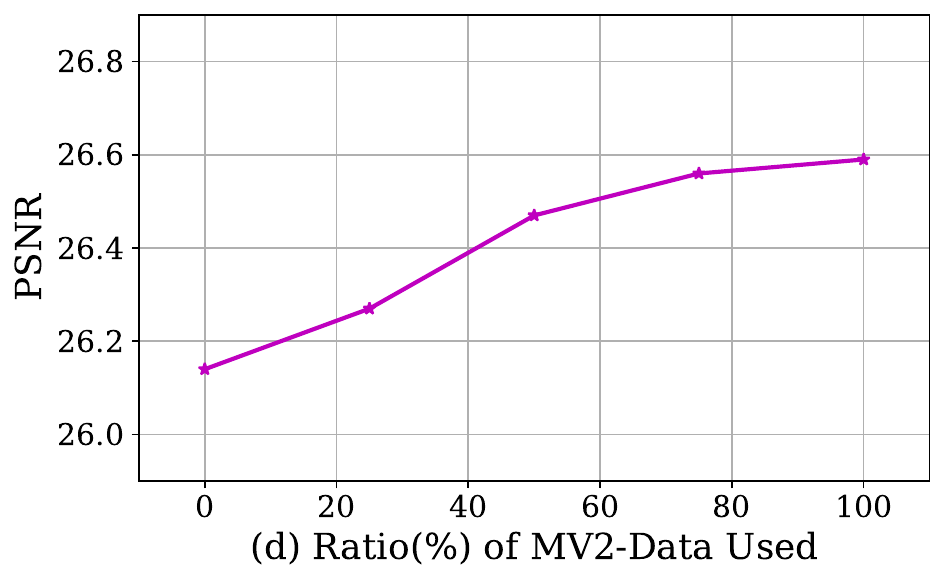} \\
	\end{tabular}
    \vspace{-0.4cm}
	\caption{The experiments for analyzing the effects of three factors in training data to the learning of LGM, including data scale (a), category range (b), and view range (c).}
	\label{fig:exp_factors}
 \vspace{-0.3cm}
\end{figure}

\section{Conclusion}
\label{sec:conclusion}
In this paper, we propose MVImgNet2.0, a larger-scale dataset with multi-view images for generic real-world objects. It expands the MVImgNet dataset and doubles the original scale and category range. Besides, MVImgNet2.0 also advances the data processing approaches to provide annotations of higher quality, including the foreground object masks, the estimated camera poses, and the reconstructed point clouds. To validate the value of MVImgNet2.0 data, we conduct extensive experiments on the task of 3D reconstruction and demonstrate that MVImgNet2.0 data not only can be used as high-quality per-scene reconstruction data but also is promising to provide a stronger 3D prior for generalizable object reconstruction. All data and annotations will be released to the public to inspire the computer vision and graphics communities.
\paragraph{Limitations and future work.} MVImgNet2.0, while impressive, does have some limitations. Firstly, due to the challenges in capturing, we have excluded large-scale objects such as buildings and dynamic subjects like animals that are difficult to stabilize for imaging. Future multi-view datasets could aim to capture these ``hard categories'' to facilitate learning a more comprehensive 3D prior. Secondly, there is significant potential for improvement in annotation quality, particularly in dense reconstructions, which is an important area for future work, especially with the ongoing advancements in dense reconstruction techniques. Thirdly, the majority of the videos focus on a single central object, resulting in camera trajectories that are relatively monotonous, \textit{i.e.} orbiting around the main subject. We plan to expand our collection to include videos with complex arrangements of multiple objects and to explore alternative camera trajectory modes to support more sophisticated scene-level reconstructions.

\section*{Acknowledgements}

\label{sec:acknowledgements}

The work was supported in part by the Basic Research Project No. HZQB-KCZYZ-2021067 of Hetao Shenzhen-HK S\&T Cooperation Zone, Guangdong Provincial Outstanding Youth Fund(No. 2023B1515020055), NSFC-61931024, and Shenzhen Science and Technology Program No. JCYJ20220530143604010. It is also partly supported by  the National Key R\&D Program of China with grant No. 2018YFB1800800, by Shenzhen Outstanding Talents Training Fund 202002, by Guangdong Research Projects No. 2017ZT07X152 and No. 2019CX01X104,by Key Area R\&D Program of Guangdong Province (Grant No. 2018B030338001), by the Guangdong Provincial Key Laboratory of Future Networks of Intelligence (Grant No. 2022B1212010001), and by Shenzhen Key Laboratory of Big Data and Artificial Intelligence (Grant No. ZDSYS201707251409055).

\clearpage

\begin{figure*}[t]
	\centering
        \includegraphics[width=0.92\linewidth]{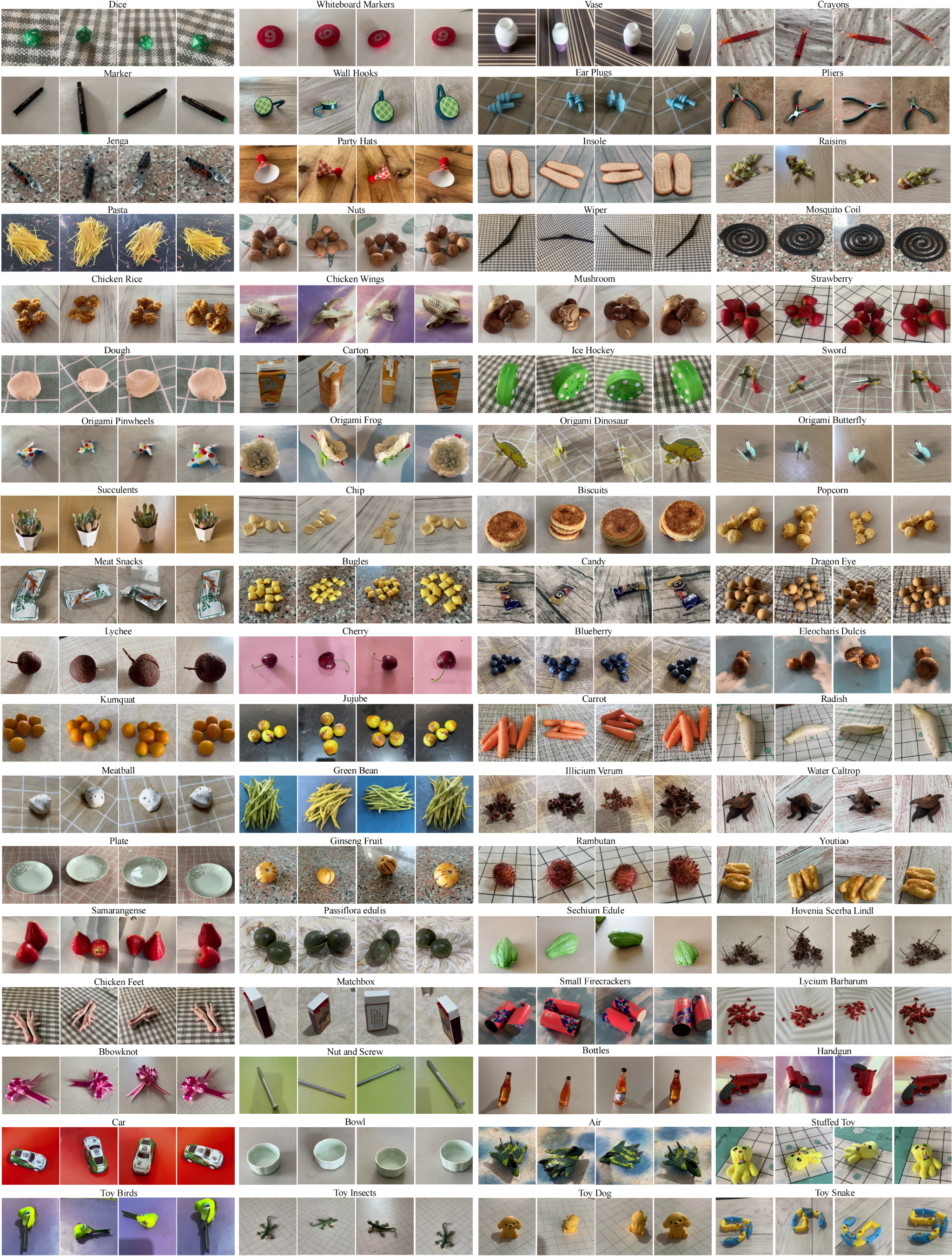}
    \vspace{-0.4cm}
	\caption{Visualization of more multi-view data from different object categories in MVImgNet2.0. }
	\label{fig:supp_mvimg4view}
 \vspace{-0.4cm}
\end{figure*}

\clearpage

\bibliographystyle{ACM-Reference-Format}
\bibliography{reference}

\clearpage
\appendix

\renewcommand{\thefigure}{R.\arabic{figure}}
\renewcommand{\thetable}{R.\arabic{table}}
\renewcommand{\theequation}{S.\arabic{equation}}

\setcounter{figure}{0}
\setcounter{table}{0}
\setcounter{equation}{0}



\section{More Details about MVImgNet2.0 Data}

\begin{figure}[b]
	\centering
	\includegraphics[width=0.9\linewidth]{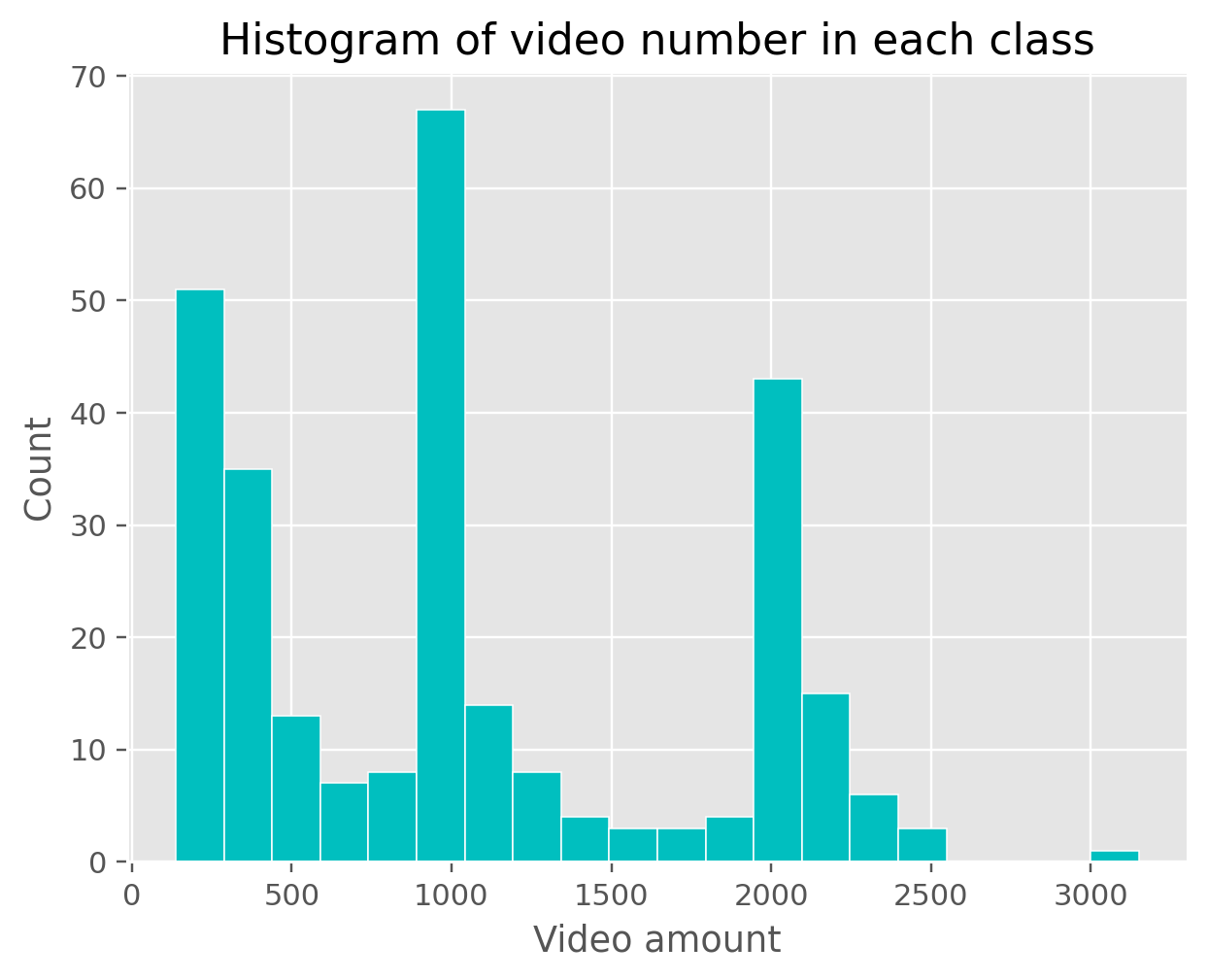}
    \vspace{-0.3cm}
	\caption{The histogram of object number in each class. }
	\label{fig:supp_hist_count}
 \vspace{-0.4cm}
\end{figure}

\begin{figure*}[t]
	\centering
	\begin{tabular}{ccc}
		\insertimg{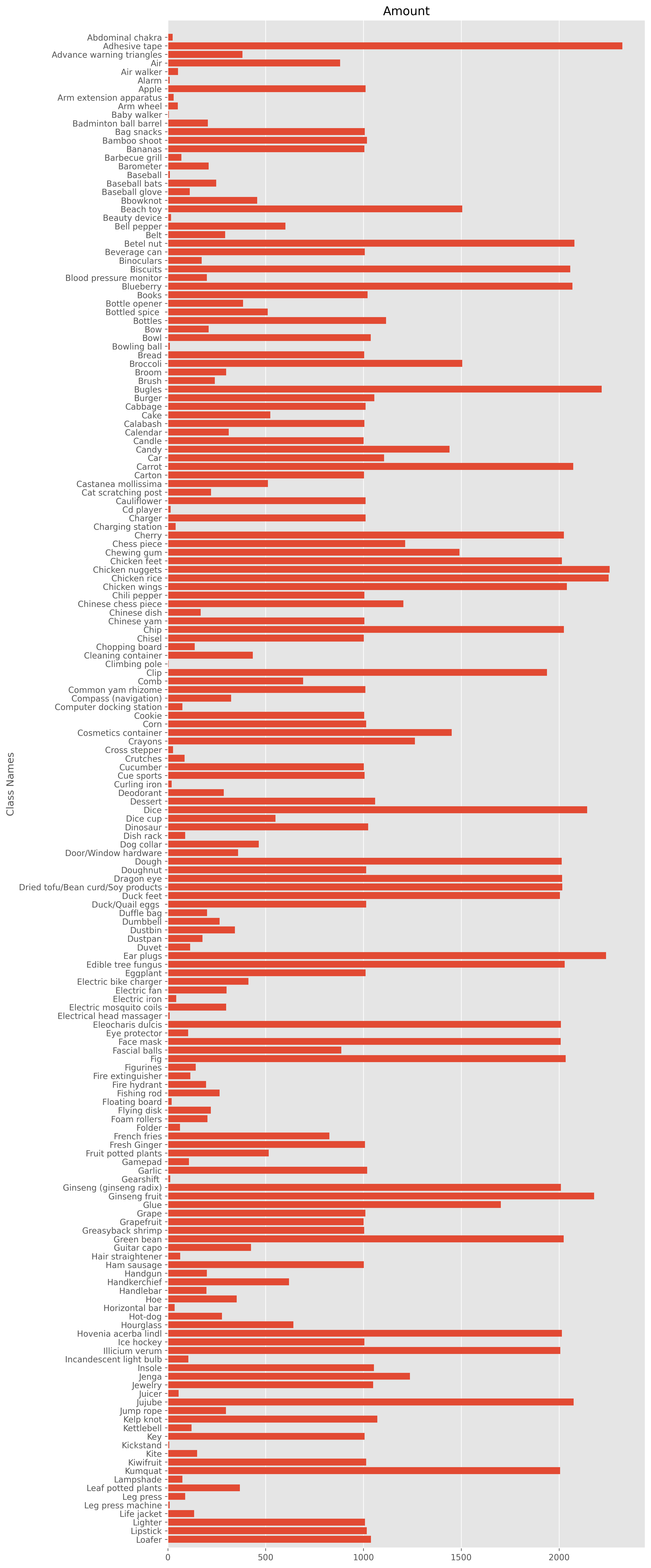}
		\insertimg{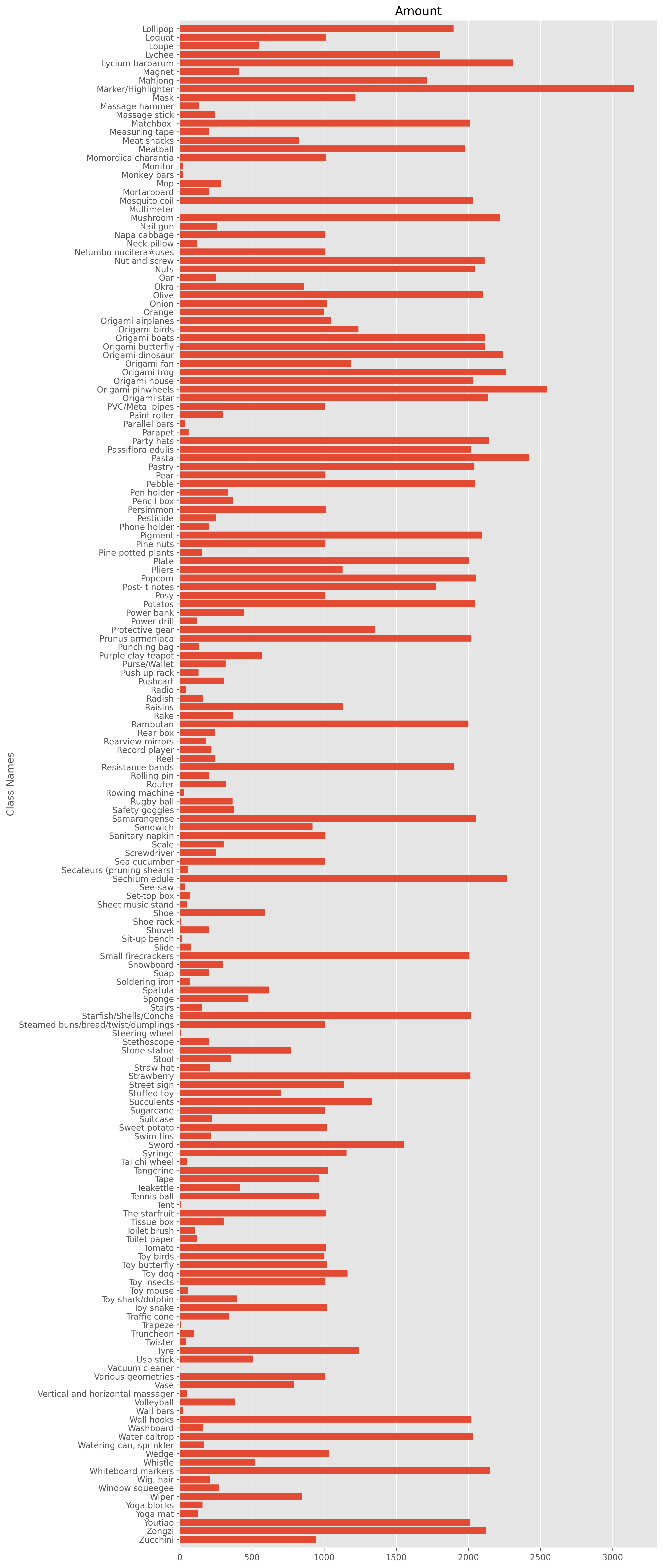}
	\end{tabular}
    \vspace{-0.3cm}
	\caption{Amounts of objects in each category in the proposed MVImgNet2.0 dataset.}
	\label{fig:supp_cls_amount}
 \vspace{-0.4cm}
\end{figure*}

\paragraph{Per-category data distributions} We count the number of object videos in each category in the proposed MVImgNet2.0 dataset. Note that among all 347 classes, 70 are old classes from MVImgNet~\cite{yu2023mvimgnet}, and these classes are not required to collect more than 1000 videos in the data acquisition process of MVImgNet2.0. Excluding them, $\sim$60\% of classes (164/277) cover 1000 or more objects. We provide a histogram for the number of objects in each MVImgNet2.0 new class in Fig.~\ref{fig:supp_hist_count}. As shown, categories can be divided into three groups: the first group is ``hard classes'', where less than 500 videos can be collected, while another two groups of categories can get around 1000 and even 2000 videos collected, respectively. A more detailed statistics of the number of objects in each category is shown in Fig.~\ref{fig:supp_cls_amount}.

\begin{figure*}[t]
	\centering
	\includegraphics[width=0.98\linewidth]{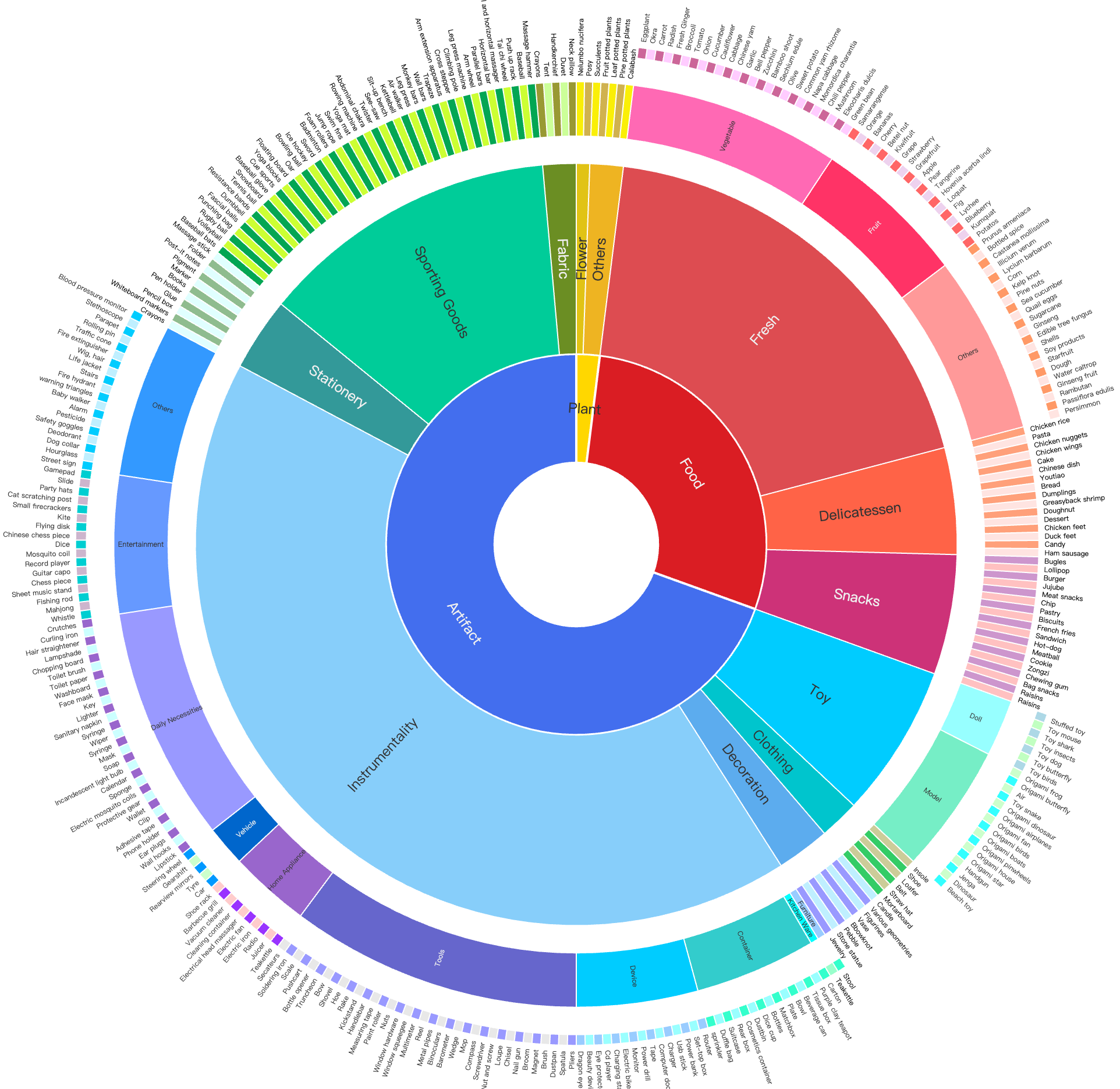}
    \vspace{-0.3cm}
	\caption{The category taxonomy of the proposed MVImgNet2.0 dataset. }
	\label{fig:supp_taxonomy}
 \vspace{-0.4cm}
\end{figure*}

\paragraph{Category taxonomy} 
The category taxonomy is shown in Fig.~\ref{fig:supp_taxonomy} to better exhibit the categories and their hierarchical relationships in MVImgNet2.0. 
\begin{table}[h]
	\centering
    \caption{Quantitative segmentation results (MSE$\downarrow$ $\times10^{-1}$) on the ECCSD dataset, the DAVIS dataset, and a subset of 500 MVImgNet images (MV1-500) with ground-truth object masks. } 
    \vspace{-3mm}
    \resizebox{0.75\linewidth}{!}{
        \begin{tabular}{l|ccc}
        \toprule
        Methods & ECCSD & DAVIS & MV1-500 \\ 
        \midrule
        MV1-Anno          & 0.143 & 0.195 & 0.243 \\
        MV2-Anno (ours)   & 0.103 & 0.143 & 0.172\\
        \bottomrule
        \end{tabular}
    }  
	\label{tab:supp_exp_mask} 
\vspace{-0.1cm}
\end{table}

\section{More Experiments}


\paragraph{Mask annotation quality.} 
In the annotation process, we adopt a detection-segmentation-tracking pipeline to generate the foreground object mask in each view. To better demonstrate the superiority of the used segmentation manner over the original one in MVImgNet, we further evaluate the performance of this pipeline on a subset of MVImgNet data where 500 frames (randomly selected from different categories) are manually annotated with object segmentation masks, and also on other object-centric datasets with ground-truth object masks, \textit{i.e.}, ECSSD~\cite{ecssd} and DAVIS~\cite{davis}. The segmentation results of CarveKit~\cite{carvekit} (the original manner, denoted as MV1-Anno) and the advanced one (MV2-Anno) are presented in Tab.~\ref{tab:supp_exp_mask}. As shown, the performance of MV2-Anno surpasses MV1-Anno by a considerable margin.

\begin{figure}[h]
	\includegraphics[width=0.94\linewidth]{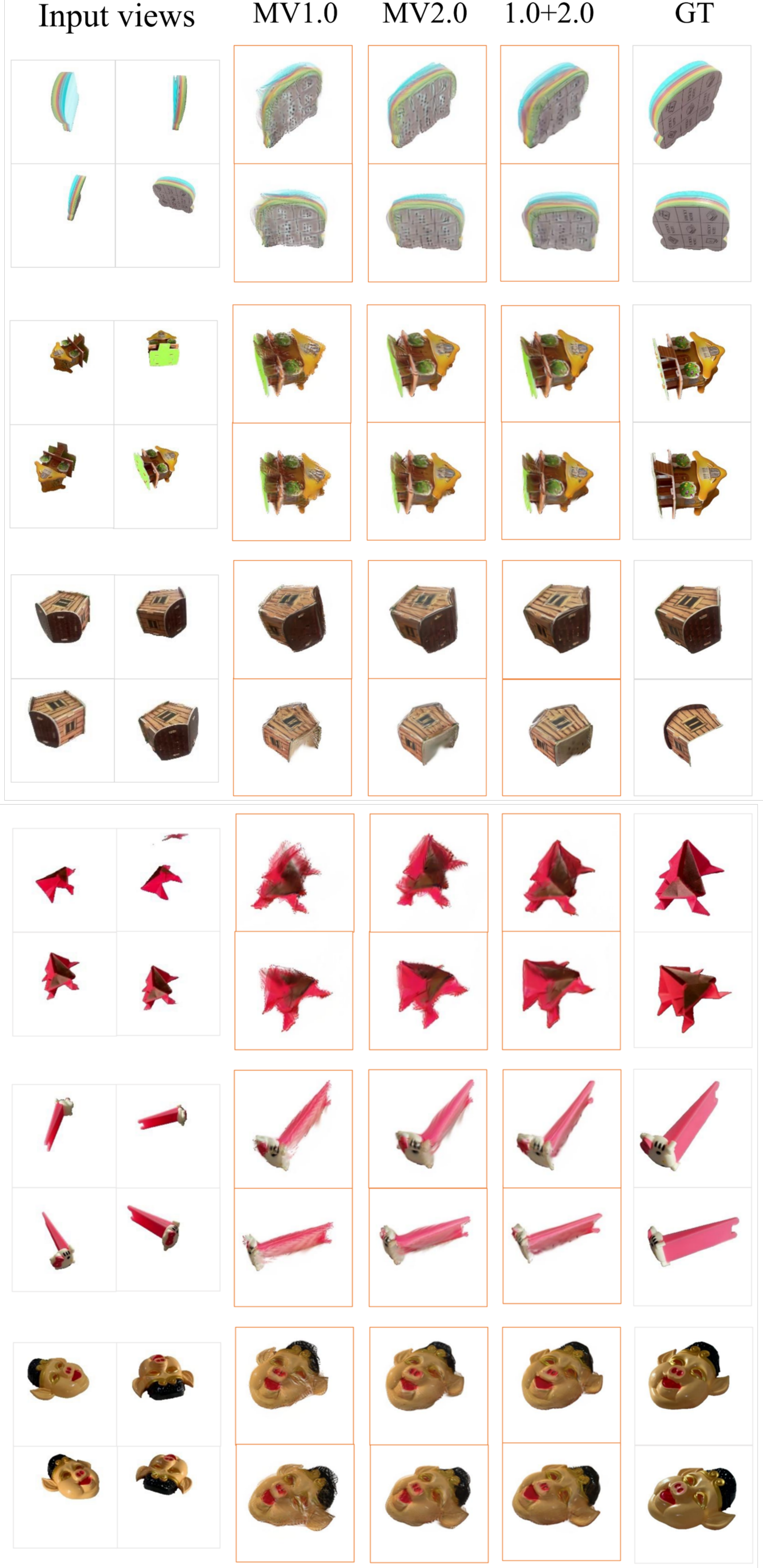}
    \vspace{-0.3cm}
	\caption{More qualitative results of LGM-tiny when trained with different kinds of data: MVImgNet1.0 (MV1.0), MVImgNet2.0 (MV2.0), and both. }
	\label{fig:supp_fig_lgm}
 \vspace{-0.4cm}
\end{figure}

\paragraph{More qualitative results.} We visualize more results of LGM-tiny for multi-view object reconstruction. We mainly show the reconstruction quality of LGM-tiny when trained with MVImgNet1.0 data (MV1-Data), MVImgNet2.0 data (MV1-Data), and both of them. As shown in Fig.~\ref{fig:supp_fig_lgm}, LGM trained with MV2-Data can achieve a higher reconstruction ability, and the use of both of them can further lead to improvements.





\end{document}